\ifdefined\XeTeXversion\else
\pdfoutput=1
\fi

\documentclass[11pt]{article}
\usepackage[final]{acl}

\usepackage{times}
\usepackage{latexsym}
\usepackage{microtype}
\usepackage{url}
\usepackage[T1]{fontenc}
\usepackage[utf8]{inputenc}

\usepackage{amsfonts}
\usepackage{amsmath}
\usepackage{amssymb}        
\usepackage{bold-extra}     
\usepackage{bm}             

\usepackage{graphicx}       
\usepackage{multirow}       
\usepackage{booktabs}       
\usepackage{enumitem}       
\usepackage{subcaption}     
\usepackage[hang,flushmargin]{footmisc}  

\title{AMARIS: A Memory-Augmented Rubric Improvement System for Rubric-Based Reinforcement Learning}

\author{Peilin Wu\textsuperscript{1}, Xinlu Zhang\textsuperscript{3}, Kun Wan\textsuperscript{2}, Wentian Zhao\textsuperscript{2}, Gang Wu\textsuperscript{2}, Xinya Du\textsuperscript{1}, Zhiyu Chen\textsuperscript{1} \\
\textsuperscript{1}The University of Texas at Dallas, \textsuperscript{2}Adobe Inc.\\
\textsuperscript{3}Department of Computer Science, University of California, Santa Barbara,\\
\texttt{\{peilin.wu,zhiyu.chen2\}@utdallas.edu} \\
}

\begin{document}
\maketitle

\begin{abstract}
Rubric-based reward shaping provides interpretable and editable reward signals for fine-tuning LLMs via reinforcement learning (RL), but existing adaptive rubric methods typically update criteria from local evidence such as the current batch or instance-level comparisons. This local view discards diagnostic information produced during training, making it difficult to track recurring failures, evaluate previous rubric edits, or raise standards once earlier criteria become saturated. We introduce AMARIS, \textbf{A} \textbf{M}emory-\textbf{A}ugmented \textbf{R}ubric \textbf{I}mprovement \textbf{S}ystem that grounds rubric updates in longitudinal training evidence. AMARIS stores rollout analyses, step-level summaries, and rubric update records in a persistent evaluation memory, then retrieves recent and semantically relevant history to revise rubrics. We evaluate AMARIS across science, medicine, instruction following, and creative writing under both global and instance-specific rubric settings. AMARIS improves over static, local-adaptive, and memory-ablated baselines, such as +2.8 points on GPQA-Diamond and +2.2 points on IFBench over the strongest baselines, while analysis shows that memory reduces oscillatory rubric edits and supports a progression from early failure correction to later curriculum advancement. AMARIS runs asynchronously alongside the normal RL loop, reducing blocking latency relative to synchronous rubric updates. 
The source code of AMARIS is at \url{https://github.com/qualidea1217/AMARIS}.
\end{abstract}

\begin{figure*}[htbp!]
\centering
\includegraphics[width=\textwidth]{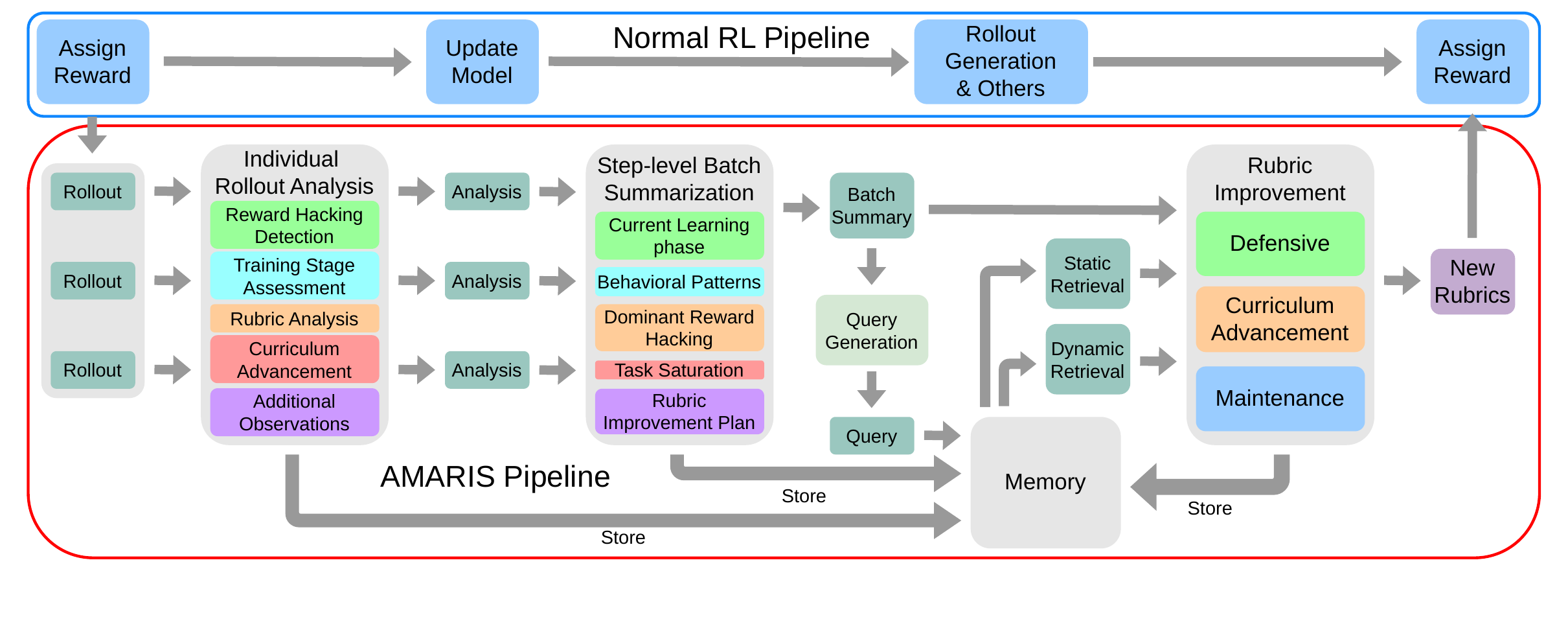}
\caption{The AMARIS system operates asynchronously in parallel with the normal RL pipeline to continuously refine rubrics.}
\label{fig:amaris}
\end{figure*}

\section{Introduction}
\label{sec:introduction}

Rubric-based reward shaping has become a practical mechanism for training language models in domains where output quality is multi-dimensional. Rubric-based approaches construct reward signals by decomposing evaluation into explicit natural-language criteria with associated weights \citep{kim2024prometheusinducingfinegrainedevaluation,ye-etal-2025-improving}. These criteria can cover correctness, helpfulness, safety, style, and instruction following, making the reward function interpretable and editable in open-ended settings where a single ground-truth verifier is hard to define \citep{10.5555/3666122.3668142}. Rubrics also allow practitioners to inspect which behaviors are being rewarded and revise the reward specification when training exposes new failure modes, including shortcuts that raise reward without improving output quality \citep{10.5555/3600270.3600957,10.5555/3618408.3618845}. However, a fixed rubric set defined before training cannot anticipate how policy behavior changes during RL training, motivating adaptive rubric updates.

Recent work improves rubric flexibility in several ways. Some methods generate criteria for each input \citep{gupta-etal-2025-carmo} or let the policy model propose rubrics for its own outputs \citep{sheng2026reinforcingchainofthoughtreasoningselfevolving,xu2026alternatingreinforcementlearningrubricbased}. Others update rubrics from pairwise comparisons between current and reference rollouts \citep{rezaei2025onlinerubricselicitationpairwise}, joint optimization with the judge \citep{xu2026alternatingreinforcementlearningrubricbased}, or batch-level selection of criteria that best distinguish current rollouts \citep{shao2025drtulureinforcementlearning}. These approaches show that rubrics can be adapted rather than fixed before training. However, most adaptive rubric updates rely on local signals such as the current instance or immediate comparisons between instances, while diagnostic evidence from earlier training steps is discarded after immediate use. As a result, each update has limited access to how policy behavior has changed over time or how previous rubric edits affected the reward signal. Reusing this evidence can make rubric adaptation more consistent across training steps and can support updates that reflect both current rollout quality and earlier evaluation history \citep{li2025curriculumrlaifcurriculumalignmentreinforcement,freitag2025curriculumreinforcementlearningcomplex}.

We introduce AMARIS, \textbf{A} \textbf{M}emory-\textbf{A}ugmented \textbf{R}ubric \textbf{I}mprovement \textbf{S}ystem for adaptive rubric-based reinforcement learning. AMARIS stores rollout-level diagnostic reports, step-level summaries, and rubric update records in a persistent evaluation memory. At each RL update step, it retrieves both recent summaries and semantically relevant historical evidence, then updates the active rubric set through defensive correction, curriculum advancement, or maintenance. This design turns rubric adaptation from a local per-step heuristic into a longitudinal evaluation process grounded in accumulated training evidence.

We evaluate AMARIS across science, medicine, instruction following, and creative writing under both global and per-instance rubric settings. AMARIS improves over static, local-adaptive, and memory-ablated baselines, including +2.8 points on GPQA-Diamond, +3.0 points on HealthBench, and +2.2 points on IFBench over the strongest non-AMARIS baselines. Ablations show that recent-step retrieval and semantic retrieval provide complementary gains, while latency analysis shows that asynchronous execution substantially reduces wall-clock delay relative to synchronous rubric updates. Analyses of rubric evolution and qualitative cases further show how memory supports defensive correction, curriculum advancement, and late-stage maintenance.

Our contributions are as follows:
\begin{itemize}
    \item We identify persistent evaluation memory as a useful state variable for adaptive rubric-based reward shaping.
    \item We propose AMARIS, a memory-augmented rubric improvement system that stores rollout analyses, step-level summaries, and rubric update records, and retrieves both recent and semantically matched history to guide rubric updates.
    \item We evaluate AMARIS under global and per-instance rubric settings across four domains and analyze rubric evolution, stability, and pipeline latency, showing how memory supports defensive correction, curriculum advancement, and maintenance.
\end{itemize}

\section{AMARIS}
\label{sec:method}

AMARIS introduces a memory-augmented rubric improvement loop into rubric-based reinforcement learning. The procedure has three stages. First, AMARIS analyzes individual rollouts as described in Section \ref{ssec:analysis}. Second, it summarizes rollout analyses into a step-level batch summary as described in Section \ref{ssec:summarization}. Third, it improves rubrics by grounding every change in recent and semantically matched analyses retrieved from memory as described in Section \ref{ssec:update}. Section \ref{ssec:pipeline} describes how this procedure can be executed asynchronously and in parallel with the normal RL pipeline.

\subsection{Individual Rollout Analysis}
\label{ssec:analysis}
In a standard rubric-based RL paradigm, the policy model's rollouts $y \sim \pi_\phi(\cdot \mid x)$, where $\pi_\phi$ is the policy model, $x$ is the input, and $y$ is the output, are evaluated against a set of active rubrics. At each RL training step $t$, AMARIS maintains an active rubric set $G_t = \{(g_j,w_j)\}_{j=1}^{n_{g,t}}$ where $g_j$ is the natural language definition of each individual rubric, $w_j$ is the scalar weight, and $n_{g,t}$ is the number of active rubrics. This rubric set can be global, with one set and one evaluation memory shared across all inputs, or per-instance, with each unique input maintaining its own set and memory. We use $G_t$ for both rubric scopes in the main text. In the per-instance setting, it denotes the active rubric set associated with the current input. The final scalar reward $r_t$ is then generated as $r_t(x,y) = S_\theta(x,y,G_t)$, where $S_\theta$ is the process of scoring using an LLM as a judge (prompt template in Figure \ref{fig:prompt_scorer_1}-\ref{fig:prompt_scorer_2}). The scalar reward is the scoring signal used by RL. Since scalar rewards may hide model behaviors such as reward hacking and rubric limitations, AMARIS also performs a deeper diagnosis of each individual rollout about the model's behavior and how the current rubrics shape it. The later rubric update stage uses batch summaries and retrieved memory to produce the next active rubric set.

For each rollout, AMARIS receives the model's input and output, the current rubric set, training metadata (the high-level goal, current step, total steps, model size, etc.), and supplemental context such as ground-truth answers, denoted by $z$, and produces a structured diagnostic report $a = A_\psi(x,y,G_t,z)$, where $A_\psi$ is the LLM for analysis (prompt template in Figures \ref{fig:prompt_analysis_1}-\ref{fig:prompt_analysis_4}). Each analysis provides five types of information. (i) \textbf{Reward hacking detection} reports whether the rollout shows signs of reward hacking, such as over-refusal for safety or sycophancy, with evidence and confidence level. (ii) \textbf{Training stage assessment} judges whether the observed performance is expected given the model size at the current step. (iii) \textbf{Rubric analysis} explains how the current rubrics may have shaped the observed behavior and whether there are specific flaws in rubrics or weights that incentivize undesired outputs. (iv) \textbf{Curriculum advancement} identifies which capability or behavior should be prioritized next to advance the model toward the high-level training goal. (v) \textbf{Additional observations} preserve other noteworthy signals that may be relevant for later analysis. The resulting structured analysis is stored in the persistent evaluation memory, preserving the evidence needed for step-level summarization in Section \ref{ssec:summarization} and rubric improvement in Section \ref{ssec:update}.

\subsection{Step-Level Batch Summarization}
\label{ssec:summarization}
Individual rollout analyses do not capture the overall step-level behavior and are too local to justify reliable rubric changes on their own. AMARIS therefore aggregates the individual analyses collected for the update at step $t$ into a step-level summary about what the model is currently doing, where it is failing, and whether the current rubrics remain informative.

Once all the analyses $\{a_j\}_{j=1}^{n_{a,t}}$ for individual rollouts are produced, where $n_{a,t}$ is the number of analyzed rollouts for the update at step $t$, AMARIS summarizes them into a step-level batch summary $b_t = B_\psi(\{a_j\}_{j=1}^{n_{a,t}})$, where $B_\psi$ is the LLM for batch summarization (prompt template in Figures \ref{fig:prompt_summary_1}-\ref{fig:prompt_summary_4}). The summary contains the general health of training and the current learning phase, the behavioral patterns that recur throughout the batch, the dominant reward hacking risk and its evidence, the weakest rubric, signs of task saturation and a provisional plan for the rubric improvement. By synthesizing step-level trends, $b_t$ provides the rubric improvement stage with a view of what the policy is currently doing and where the current set of rubrics fall short.

When an update contains more analyses than can be processed in a single summarization pass, AMARIS summarizes smaller chunks independently and consolidates the intermediate summaries into the same final step-level summary format. This hierarchical strategy, referred to as meta summarization, allows the pipeline to scale to large rollout batches while preserving a fixed downstream interface for the retrieval and rubric improvement stages. The step-level batch summaries are also stored in the persistent evaluation memory for the future rubric improvement stage in Section \ref{ssec:update}.

\subsection{Memory-Augmented Rubric Improvement}
\label{ssec:update}
Without memory, rubric updates are reactive and can respond only to the changes for the current step, with no mechanism for recalling whether a similar suboptimal behavior appeared earlier, a previous rubric update failed, or the model has been steadily improving. AMARIS addresses this limitation through persistent memory together with a hybrid retrieval strategy.

\paragraph{Persistent evaluation memory.}
AMARIS stores all previous evaluations in a database that persists throughout training as the memory available at step $t$, denoted by $M_t$. The memory continuously accumulates individual rollout analyses $a$, step-level summaries $b$, chunk-level summaries if meta summarization is enabled, and rubric improvement records $u$. Here, $u$ is a structured record containing the rubric update strategy, retrieved history consulted, rubric edit operations, resulting adaptive rubric set, and a short summary for future steps. Each evaluation document is indexed for later retrieval. The memory therefore contains rich information about the past evaluations and supports various retrieval strategies.

\paragraph{Static and dynamic retrieval.}
Before each update, AMARIS retrieves relevant historical context $I_t$ from static and dynamic retrieval. Static retrieval supplies recent step-level summaries, giving the updater a compact view of the recent training trajectory with temporal continuity. Dynamic retrieval reads the current step-level summary and uses the query generator $Q_\psi$ to generate targeted search queries (prompt template in Figure \ref{fig:prompt_query_1}-\ref{fig:prompt_query_3}). These queries usually cover the dominant failure mode, suspected rubric flaw, possible curriculum objective, and failure from similar past updates. The retrieved documents may include earlier individual analyses, summaries, and rubric update records from any point in training. Duplicate documents are removed before rubric update. This retrieval policy keeps the updater context bounded while maintaining flexibility.

\paragraph{Rubric improvement.}
Based on the retrieval results, AMARIS generates the next set of rubrics as $G_{t+1}, u_t = U_\psi(G_t,b_t,I_t)$, where $U_\psi$ is the LLM for rubric improvement (prompt template in Figures \ref{fig:prompt_update_1}-\ref{fig:prompt_update_4}). AMARIS selects among three broad strategies. A \textbf{Defensive} update modifies rubrics and weights to reduce suboptimal behaviors and uses retrieved memory to check whether the same suboptimal behavior has appeared before. A \textbf{Curriculum advancement} update raises the evaluation standard when the current rubrics appear saturated and retrieved memory confirms that the model has mastered the easier requirements. A \textbf{Maintenance} update keeps the rubric set stable when the current evidence supports stability. To realize its chosen strategy, the LLM for rubric update $U_\psi$ may apply any combination of rubric operations. \textbf{Create} adds new rubrics, \textbf{Update} revises rubric text for precision, \textbf{Delete} removes useless or counterproductive rubrics, \textbf{Reweight} shifts training emphasis, \textbf{Merge} consolidates overlapping rubrics, and \textbf{Split} decomposes one rubric into atomic rubrics. $U_\psi$ also produces reasoning summarizing what history was consulted and what lessons should be retained. The resulting update record is written back to the memory so that every update becomes available as evidence for subsequent updates.

\subsection{RL Pipeline Integration}
\label{ssec:pipeline}
The rubric improvement procedure would add substantial blocking latency if executed synchronously within each RL step. To avoid blocking the RL training loop, AMARIS executes the procedure asynchronously and in parallel with the normal RL pipeline. Specifically, for every step, rollouts are scored immediately using the currently active set of rubrics, while the rubric improvement proceeds in the background. Once the update completes, the updated rubrics become active for subsequent scoring. Although such asynchronous execution causes the rubrics to be slightly off-policy, it reduces blocking latency by moving rubric improvement off the critical path. It also requires less interference with the original RL training process and fewer changes to the code of the RL training pipeline.

\section{Experiment Setup}
\label{sec:experiment_setup}
This section describes the experimental framework used to evaluate AMARIS. We outline the datasets, evaluation benchmarks, baselines, and implementation details to provide a clear context for our results.

\subsection{Datasets \& Evaluation}
\label{ssec:datasets_and_eval}
We evaluate AMARIS across four diverse domains with both open-ended and closed-ended tasks. For each domain, we pair a rubric-annotated training set with one or more evaluation benchmarks. (i) \textbf{Science.} We train on RaR-Science \citep{gunjal2025rubricsrewardsreinforcementlearning}, a corpus of scientific questions annotated with rubrics, and evaluate on GPQA-Diamond \citep{rein2024gpqa}. We report average accuracy (\%) over 10 measurements as the metric. (ii) \textbf{Medicine.} We train on RaR-Medicine \citep{gunjal2025rubricsrewardsreinforcementlearning}, a medical variant of the rubric-annotated training data, and evaluate on HealthBench \citep{arora2025healthbenchevaluatinglargelanguage}. We report the HealthBench composite score as the metric. (iii) \textbf{Instruction following.} We follow the experimental setup of OpenRubrics \citep{liu2026openrubricsscalablesyntheticrubric} and train on the OpenRubrics \citep{liu2026openrubricsscalablesyntheticrubric} dataset. We evaluate on three instruction-following benchmarks: IFEval \citep{zhou2023instructionfollowingevaluationlargelanguage}, InfoBench \citep{qin-etal-2024-infobench}, and IFBench \citep{pyatkin2025generalizingverifiableinstructionfollowing}, which we report the overall accuracies of each of them. (iv) \textbf{Creative writing.} We train on the writing subset of RubricHub \citep{li2026rubrichubcomprehensivehighlydiscriminative}, a large-scale rubric-annotated dataset. We evaluate on WritingBench \citep{wu2025writingbenchcomprehensivebenchmarkgenerative} and Creative Writing v3 \citep{creative-writing-bench-v3} from EQ-Bench and report the overall scores as the metric, also referred as CW-v3 in the following sections.

\subsection{Baselines}
\label{ssec:baselines}
We compare AMARIS against several baselines. Three baselines apply across all four domains: (1) Naive (no RL): the base model without any RL fine-tuning. (2) RubricHub \citep{li2026rubrichubcomprehensivehighlydiscriminative}: a large scale rubric-annotated dataset with evaluations on rubric-based reinforcement learning (RuRL) that covers all the domains we tested. (3) RuscaRL \citep{zhou2026breakingexplorationbottleneckrubricscaffolded}: a rubric-based RL method that we evaluate across all four domains. We also include domain-specific baselines. (i) \textbf{Science \& Medicine.} We include RaR \citep{gunjal2025rubricsrewardsreinforcementlearning}, which uses a fixed set of rubrics throughout training on the same RaR-Science and RaR-Medicine data. (ii) \textbf{Instruction following.} We include OpenRubrics \citep{liu2026openrubricsscalablesyntheticrubric}, Rubric-ARM \citep{xu2026alternatingreinforcementlearningrubricbased}, and RLCF \citep{viswanathan2025checklistsbetterrewardmodels}. For all the baselines, we report the values obtained by re-training and re-evaluating using the same base model as ours, if possible. If not, we reference their reported values.

\subsection{Implementation Details}
\label{ssec:implementation_details}
Unless otherwise stated, we perform RL training with Qwen2.5-7B-Instruct \citep{qwen2025qwen25technicalreport} as the base model, GRPO \citep{shao2024deepseekmathpushinglimitsmathematical} as the RL algorithm with group-wise and batch-wise reward normalization enabled, using 4 NVIDIA A100 80GB GPUs. We train all models with up to 400 steps and save the checkpoint every 50 steps, and use the final saved checkpoint for testing to ensure a fair evaluation. By default, we use GPT-OSS-120B \citep{openai2025gptoss120bgptoss20bmodel} for scoring and individual analysis, and GPT-5.1 \citep{singh2025openaigpt5card} for the remaining usages. This allocation reflects that scoring and individual analysis account for the majority of token consumption and therefore benefit most from a cost-effective model, while the remaining stages consume fewer tokens but demand higher reasoning capability. We analyze the influence of LLM choices in Appendix \ref{sec:influence_of_underlying_llm}, and test an alternative Qwen3-4B policy base in Appendix~\ref{ssec:alternative_base_model}. The memory is implemented with Chroma \citep{chroma_repo} as the vector database. For reward computation, following previous works \citep{gunjal2025rubricsrewardsreinforcementlearning, li2026rubrichubcomprehensivehighlydiscriminative}, we use a single LLM call per rollout. For all the LLM usages including the policy and AMARIS, we use the default setting for inferences.

Unless otherwise stated, we evaluate both rubric scopes introduced in Section \ref{sec:method}. In the global setting, AMARIS initializes the shared rubric set using the same pipeline in Section \ref{sec:method} with an empty memory and the first-step batch. The first-step rollouts are then scored synchronously using the initialized rubric set. In the per-instance setting, each input-specific rubric set is initialized using the annotated rubrics provided by the dataset. For global rubrics, rubric improvement is triggered at every step by default. For per-instance rubrics, updates occur when the corresponding input reappears and new rollouts are available for analysis, since different inputs are sampled at different steps. To increase revisit frequency, we set the batch size to 256 and the number of samples per query to 4. We enable meta summarization for global rubrics with chunk size 32. We set the maximum rubric staleness to 1 version, meaning that each step scores rollouts using rubrics updated from at most one version prior. If the background pipeline has not yet completed, the training loop waits for its result before proceeding. We additionally measure performance under different rubric update intervals in Section \ref{ssec:influence_of_update_scheduling} and profile the latency in Section \ref{ssec:latency_analysis}.

\begin{table*}[htbp!]
\centering\small{
\newcommand{\meanstd}[2]{\begin{tabular}[c]{@{}c@{}}#1\\[-1pt]{\scriptsize$\pm$#2}\end{tabular}}
\newcommand{\bestmeanstd}[2]{\begin{tabular}[c]{@{}c@{}}\textbf{#1}\\[-1pt]{\scriptsize$\pm$#2}\end{tabular}}
\resizebox{\textwidth}{!}{
\begin{tabular}{l c c ccc cc}
\toprule
\multirow{2}{*}{\textbf{Method}} &
\textbf{Science} & \textbf{Medicine} &
\multicolumn{3}{c}{\textbf{Instruction Following}} &
\multicolumn{2}{c}{\textbf{Creative Writing}} \\
\cmidrule(lr){2-2} \cmidrule(lr){3-3} \cmidrule(lr){4-6} \cmidrule(lr){7-8}
& \textbf{GPQA-D} & \textbf{HealthBench} &
\textbf{IFEval} & \textbf{InfoBench} & \textbf{IFBench} &
\textbf{WritingBench} & \textbf{CW-v3} \\
\midrule
\multicolumn{8}{l}{\textit{Cross-domain baselines}} \\
Naive (no RL)                    & 35.0 & 22.7 & 77.3 & 78.1 & 28.2 & 45.2 & 37.4 \\
RubricHub (RuRL)                 & 38.8 & 33.0 & 79.8 & 83.5 & 33.5 & 56.9 & 39.0 \\
RuscaRL                          & 38.5 & 32.9 & 79.0 & 83.2 & 33.0 & 56.1 & 38.6 \\
\midrule
\multicolumn{8}{l}{\textit{Science \& Medicine baselines}} \\
RaR                              & 37.6 & 31.2 & --   & --   & --   & --   & -- \\
\midrule
\multicolumn{8}{l}{\textit{Instruction-following baselines}} \\
OpenRubrics (DPO via Rubric-RM)  & --   & --   & 79.5 & 83.0 & 33.7 & --   & -- \\
Rubric-ARM                       & --   & --   & 80.4 & 83.7 & 35.0 & --   & -- \\
RLCF                             & --   & --   & 78.6 & 84.1 & 28.2 & --   & -- \\
\midrule
\multicolumn{8}{l}{\textbf{AMARIS (Our proposed system)}} \\
AMARIS (global rubric)           & \meanstd{41.1}{0.95} & \meanstd{35.6}{1.18} & \meanstd{81.5}{0.38} & \meanstd{86.6}{0.26} & \meanstd{36.4}{0.58} & \bestmeanstd{60.4}{0.97} & \meanstd{41.7}{1.08} \\
AMARIS (per-instance rubric)     & \bestmeanstd{41.6}{0.39} & \bestmeanstd{36.0}{1.06} & \bestmeanstd{81.9}{0.44} & \bestmeanstd{87.0}{0.23} & \bestmeanstd{37.2}{0.66} & \meanstd{59.8}{0.80} & \bestmeanstd{42.3}{1.48} \\
\bottomrule
\end{tabular}}
}
\caption{Main results across four evaluation domains. Best results in each column are in \textbf{bold}. ``--'' indicates that the result is not applicable to that domain. Results reported with standard deviations are computed over five repeated evaluations of the same trained checkpoint.}
\label{tab:main_results}
\end{table*}

\section{Results \& Analysis}
This section evaluates AMARIS from four angles. Section~\ref{ssec:main_results} reports downstream results across science, medicine, instruction following, and creative writing. Section~\ref{ssec:ablation_studies_on_memory_usage} ablates memory configuration and retrieval budgets. Section~\ref{ssec:static_rubric_spectrum_main} tests whether online rubric adaptation is needed beyond static initial or final rubric sets. Section~\ref{ssec:latency_analysis} profiles the pipeline latency and compares synchronous and asynchronous execution. Additional per-instance results, LLM-component ablations, update-frequency controls, scalar-judge baselines, alternative-base-model results, rubric-quality evaluation, rubric-evolution analysis, and case studies appear in the appendix.

\subsection{Main Results}
\label{ssec:main_results}
We compare AMARIS against general and domain-specific baselines across four evaluation settings. As shown in Table~\ref{tab:main_results}, AMARIS consistently achieves the best performance on all benchmarks under both rubric strategies, with the per-instance rubric setting yielding the strongest results on six of the seven benchmarks. Relative to the strongest non-AMARIS baseline, the best AMARIS result improves by +2.8 on GPQA-Diamond, +3.0 on HealthBench, +1.5 on IFEval, +2.9 on InfoBench, +2.2 on IFBench, +3.5 on WritingBench, and +3.3 on CW-v3.

\begin{table*}[t!]
\centering\small{
\setlength{\tabcolsep}{4pt}
\renewcommand{\arraystretch}{0.98}
\resizebox{\textwidth}{!}{
\begin{tabular}{ll cc ccc cc}
\toprule
\multirow{2}{*}{\textbf{Memory}} & \multirow{2}{*}{\textbf{Setting}} &
\textbf{Science} & \textbf{Medicine} &
\multicolumn{3}{c}{\textbf{Instruction Following}} &
\multicolumn{2}{c}{\textbf{Creative Writing}} \\
\cmidrule(lr){3-3} \cmidrule(lr){4-4} \cmidrule(lr){5-7} \cmidrule(lr){8-9}
& & \textbf{GPQA-D} & \textbf{HealthBench} &
\textbf{IFEval} & \textbf{InfoBench} & \textbf{IFBench} &
\textbf{WritingBench} & \textbf{CW-v3} \\
\midrule
No memory & -- 
& 39.2 & 33.8 & 80.6 & 85.3 & 35.2 & 57.3 & 41.0 \\
\midrule
\multirow{4}{*}{Static only}
& $N = 2$ 
& 39.4 & 34.4 & 80.8 & 85.6 & 35.5 & 57.9 & 41.1 \\
& $N = 4$ 
& \textbf{40.1} & 35.1 & \textbf{81.1} & 86.0 & 35.9 & \textbf{58.6} & 41.3 \\
& $N = 6$ 
& 40.0 & 34.9 & 81.0 & 85.9 & 35.8 & 58.5 & 41.3 \\
& $N = 8$ 
& \textbf{40.1} & \textbf{35.2} & \textbf{81.1} & \textbf{86.1} & \textbf{36.0} & 58.5 & \textbf{41.4} \\
\midrule
\multirow{3}{*}{Dynamic only}
& $K = 5$ 
& 39.6 & 34.4 & 80.9 & 85.9 & 35.4 & 58.0 & 41.2 \\
& $K = 10$ 
& \textbf{39.8} & \textbf{35.3} & \textbf{81.0} & \textbf{86.2} & \textbf{36.0} & 58.2 & \textbf{41.4} \\
& $K = 15$ 
& \textbf{39.8} & 34.9 & 80.9 & 86.0 & 35.8 & \textbf{58.4} & 41.3 \\
\midrule
Static + Dynamic & $N = 4, K = 10$ 
& \textbf{41.1} & \textbf{35.6} & \textbf{81.5} & \textbf{86.6} & \textbf{36.4} & \textbf{60.4} & \textbf{41.7} \\
\bottomrule
\end{tabular}}
}
\caption{Ablation on memory configuration and retrieval hyperparameters under the global rubrics setting. Best results within the static-only and dynamic-only blocks are in \textbf{bold}.}
\label{tab:ablation}
\end{table*}

\begin{table*}[t!]
\centering\small{
\resizebox{\textwidth}{!}{
\begin{tabular}{l cc ccc cc}
\toprule
\multirow{2}{*}{\textbf{Rubric regime}} &
\textbf{Science} & \textbf{Medicine} &
\multicolumn{3}{c}{\textbf{Instruction Following}} &
\multicolumn{2}{c}{\textbf{Creative Writing}} \\
\cmidrule(lr){2-2} \cmidrule(lr){3-3} \cmidrule(lr){4-6} \cmidrule(lr){7-8}
& \textbf{GPQA-D} & \textbf{HealthBench} &
\textbf{IFEval} & \textbf{InfoBench} & \textbf{IFBench} &
\textbf{WritingBench} & \textbf{CW-v3} \\
\midrule
Initial-static      & 38.9 & 33.7 & 80.6 & 85.1 & 35.2 & 57.1 & 40.8 \\
No-memory adaptive  & 39.1 & 33.9 & 80.7 & 85.2 & 35.3 & 57.4 & 40.9 \\
Final-static        & 40.9 & \textbf{35.5} & 81.3 & 86.3 & 36.1 & 59.9 & 41.7 \\
AMARIS dynamic      & \textbf{41.2} & \textbf{35.5} & \textbf{81.6} & \textbf{86.7} & \textbf{36.3} & \textbf{60.3} & \textbf{41.8} \\
\bottomrule
\end{tabular}}
}
\caption{Static rubric spectrum under global rubrics. All conditions use the same base checkpoint, training data, RL algorithm, and training budget; only the rubric update regime changes.}
\label{tab:static_rubric_spectrum_global}
\end{table*}

\subsection{Ablation Studies On Memory Usage}
\label{ssec:ablation_studies_on_memory_usage}

We conduct two sets of experiments to validate the core design choices of AMARIS about using evaluation memory: (i) we isolate the contribution of each memory type by comparing no memory, static-only, dynamic-only, and the combined configuration; and (ii) we change the static memory retrieval number $N$ and dynamic query number $K$. Results are in Table \ref{tab:ablation} for global rubrics and Table \ref{tab:ablation_per_instance} for per-instance rubrics.

\paragraph{Memory configuration.}
Both memory types show gains over the no-memory baseline, and their combination yields the best results across all benchmarks (+1.9 on GPQA-D and +1.8 on HealthBench). Notably, the two types have complementary strengths. For example, in scientific domain, static memory slightly outperforms dynamic memory (40.1 vs.\ 39.8), suggesting that recent historical evaluation steps is valuable for scientific reasoning, possibly because rubric drift and recurring errors tend to appear within a short period of steps. For HealthBench, the pattern (35.3 vs.\ 35.1) indicates that targeted semantic retrieval is more beneficial due to suboptimal behaviors recur across distant steps, unlike the science domain. 
In all cases, the combined configuration captures both temporal and semantic signals and achieves the strongest overall performance, which we adopt as the default, and the data of per-instance rubrics shows similar trend as shown in Table \ref{tab:ablation_per_instance}.
 
\paragraph{Retrieval configuration.}
For both retrieval mechanisms, we observe performance gain up to a moderate budget followed by diminishing returns. Increasing the static retrieval amount from $N{=}2$ to $N{=}4$ provides the main improvement (e.g., +0.7 on both GPQA-D and HealthBench), suggesting that four preceding summaries are sufficient to capture the recent training trajectory and prevent oscillatory rubric changes. 
Similarly, increasing the dynamic retrieval amount by increasing the query from $K{=}5$ to $K{=}10$ produces consistent improvements across all benchmarks, but $K{=}15$ provides no consistent further benefit: most metrics remain flat or decline, while WritingBench improves only slightly (+0.2) and HealthBench drops from 35.3 to 34.9. We attribute this to additional queries retrieving marginal or noisy documents. We therefore select $N{=}4$ and $K{=}10$ as defaults. For per-instance rubric settings, the trend agrees with the global rubrics setting as shown in Table \ref{tab:ablation_per_instance}.

\subsection{Static Rubric Spectrum}
\label{ssec:static_rubric_spectrum_main}
We further test whether AMARIS gains come from online rubric adaptation or simply from obtaining a better final rubric set. Table~\ref{tab:static_rubric_spectrum_global} reports the global-rubric comparison, and Appendix~\ref{ssec:static_rubric_spectrum} reports the corresponding per-instance comparison. The overall trend is initial-static $\le$ no-memory adaptive $<$ final-static $\lesssim$ AMARIS dynamic. Frozen cold-start rubrics are consistently weakest because they have not been revised with training feedback. No-memory adaptation gives only small gains over this cold-start condition, indicating that local edits alone are insufficient when the updater cannot retrieve prior diagnostics or previous update outcomes. Final-static rubrics recover most of the improvement, showing that AMARIS learns useful criteria during training, but they remain below or tied with online AMARIS. This residual gap is consistent with the stage-specific behavior observed in Appendix~\ref{sec:rubric_evolvement}: early training benefits from defensive criteria, mid-training shifts toward curriculum advancement, and late training enters maintenance. A single frozen final set cannot provide this training-stage curriculum.

\subsection{Latency Analysis}
\label{ssec:latency_analysis}
A practical concern for any auxiliary reward pipeline is the wall-clock latency added to RL training when the pipeline sits on the critical path. We therefore profile each AMARIS component and compare asynchronous execution in Section \ref{ssec:pipeline} with traditional synchronous execution, where scoring always waits for the rubric update within the same step so that the rubric stays on-policy.
 
\paragraph{Component-level profiling.}
Table \ref{tab:latency_breakdown} shows the average time and token consumption of each AMARIS component measured over one complete training using the default setting of AMARIS under global rubrics. The individual rollout analysis dominates both time (63.1\%) and tokens (2.32\,M per step), as it processes every rollout. The batch summarization is the second most expensive stage (28.5\%), while the other stages together account for less than 9\% of the total pipeline time. 
 
\begin{table}[t!]
\centering\small{
\resizebox{\columnwidth}{!}{
\begin{tabular}{l r r r}
\toprule
\textbf{Component} & \textbf{Avg.\ time (s)} & \textbf{\% of pipeline} & \textbf{Tokens / step} \\
\midrule
Individual analysis  & 426.4 & 63.1\% & 2.32\,M \\
Batch summarization     & 192.3 & 28.5\% & 246\,k \\
Query generation      &   2.7 &  0.4\% & 14.1\,k \\
Memory retrieval     &  23.7 &  3.5\% & 0 \\
Rubric update       &  30.6 &  4.5\% & 110.9\,k \\
\midrule
Total pipeline       & 675.7 & 100.0\% & 2.691\,M \\
\bottomrule
\end{tabular}}
}
\caption{Per-component time and token consumption of the AMARIS averaged over 400 training steps.}
\label{tab:latency_breakdown}
\end{table}
 
\paragraph{Asynchronous vs.\ Synchronous execution.}
\begin{table*}[t!]
\centering\small{
\resizebox{\textwidth}{!}{
\begin{tabular}{l cc ccc cc r r}
\toprule
\multirow{2}{*}{\textbf{Pipeline mode}} &
\textbf{Science} & \textbf{Medicine} &
\multicolumn{3}{c}{\textbf{Instruction Following}} &
\multicolumn{2}{c}{\textbf{Creative Writing}} &
\multirow{2}{*}{\textbf{Time (h)}} &
\multirow{2}{*}{\textbf{Time increase}} \\
\cmidrule(lr){2-2} \cmidrule(lr){3-3} \cmidrule(lr){4-6} \cmidrule(lr){7-8}
& \textbf{GPQA-D} & \textbf{HealthBench} &
\textbf{IFEval} & \textbf{InfoBench} & \textbf{IFBench} &
\textbf{WritingBench} & \textbf{CW-v3} & & \\
\midrule
Sync  & \textbf{41.3} & \textbf{35.7} & \textbf{81.6} & 86.4 & \textbf{36.6} & 60.0 & \textbf{41.8} & ${\sim}$146 & ${\sim}$92\% \\
Async & 41.1 & 35.6 & 81.5 & \textbf{86.6} & 36.4 & \textbf{60.4} & 41.7 & ${\sim}$80 & ${\sim}$5\% \\
\bottomrule
\end{tabular}}
}
\caption{Synchronous vs.\ asynchronous pipeline execution. Async mode substantially reduces wall-clock latency relative to synchronous execution while incurring only marginal quality differences ($\le$0.2 across most benchmarks).}
\label{tab:sync_async}
\end{table*}
 
Table \ref{tab:sync_async} shows the quality-latency trade-off. Synchronous execution yields mostly a 0.2-point advantage on any single benchmark (e.g., 41.3 vs.\ 41.1 on GPQA-D). This marginal gain comes at the cost of nearly doubling the total training time. Asynchronous execution therefore provides a substantially more favorable wall-clock trade-off and is adopted as the default pipeline mode. We also compare the total time consumption using both asynchronous and synchronous AMARIS with the normal rubric-based RL training with static rubrics only in terms of the Time increase column in Table \ref{tab:sync_async}. The results show that asynchronous execution sharply reduces blocking latency by moving rubric improvement off the training critical path. We also profile the latency for per-instance rubric settings and find that the additional wall-clock delay is small. This is because the per-instance setting naturally limits update opportunity but provides abundant time for analysis and rubric update, especially for summarization and updating that can be further parallelized by using external APIs.

\section{Related Work}
\label{sec:related_work}

\subsection{Rubric-based RL for LLM}
LLM alignment has evolved from scalar reward models trained on pairwise preferences \citep{10.5555/3600270.3602281,bai2022traininghelpfulharmlessassistant} toward reward signals that expose more of the underlying rubrics, making the reward specification more interpretable and editable \citep{kim2024prometheusinducingfinegrainedevaluation, gunjal2025rubricsrewardsreinforcementlearning, viswanathan2025checklistsbetterrewardmodels}. This paradigm extends RL beyond strictly verifiable domains. However, a rubric set is still a reward specification: if it is static, incomplete, or overly coarse, optimization can exploit gaps in the evaluator in the same spirit as reward model overoptimization \citep{10.5555/3618408.3618845}. This motivates methods that adapt rubrics as policy behavior changes during training.

\subsection{Adaptive Rubrics}
Adaptive rubric methods address the static-specification problem by generating or revising criteria during evaluation and training. CARMO \citep{gupta-etal-2025-carmo} and OnlineRubrics \citep{rezaei2025onlinerubricselicitationpairwise} elicit criteria from context or pairwise comparisons, while Rubric-ARM \citep{xu2026alternatingreinforcementlearningrubricbased} and DR Tulu \citep{shao2025drtulureinforcementlearning} couple policy optimization with bounded rubric refinement or evolving rubrics. Related reward-modeling work reduces reward hacking or guides learning through causal rubric structure \citep{srivastava2025robustrewardmodelingcausal}, information-theoretic regularization \citep{10.5555/3737916.3742186}, and staged AI feedback \citep{li2025curriculumrlaifcurriculumalignmentreinforcement}. These approaches make the reward specification more flexible, but most updates are grounded in local evidence such as the current input, pairwise comparisons, or a predefined curriculum. They typically do not preserve rollout diagnostics, score trends, and previous rubric-edit outcomes as reusable state for later updates.

\subsection{Memory for Longitudinal Usage}
External memory systems for LLM agents show that persistent state can support long-horizon adaptation. A-Mem \citep{xu2025amemagenticmemoryllm}, Memory-R1 \citep{yan2026memoryr1enhancinglargelanguage}, and Meta-Policy Reflexion \citep{wu2025metapolicyreflexionreusablereflective} introduce read-write-consolidation mechanisms for retaining and reusing experience across tasks, reflecting a broader shift toward memory-augmented agents \citep{hu2026memoryageaiagents}. AMARIS extends memory-augmented systems to evaluation-state maintenance, where the stored objects are rollout diagnostics, score trends, and rubric-edit outcomes rather than task-solving experiences. Rubric updates can then use recent trajectories and related past evidence instead of isolated local edits.

\section{Conclusion}
We present AMARIS, a memory-augmented rubric improvement system for rubric-based reinforcement learning. AMARIS stores rollout analyses, step-level summaries, and rubric update records in a persistent evaluation memory, and uses both static and dynamic retrieval to ground rubric changes in training history. Across science, medicine, instruction following, and creative writing domains, AMARIS improves over strong rubric-based baselines under both global and per-instance rubric settings. Ablation studies show that both retrieval modes contribute to the gains, while asynchronous execution reduces the blocking latency of rubric improvement during training.

\section*{Limitations}
AMARIS requires additional inference resources to analyze rollouts, summarize batches, retrieve historical evidence, and update rubrics. Asynchronous execution reduces the blocking latency observed by the RL training loop, but it does not reduce the total compute required by these auxiliary stages. Future work could further optimize the update interval and scheduling policy. The update-frequency study in Appendix~\ref{ssec:influence_of_update_scheduling} shows that updating every 2 steps is slightly better than updating every step on several benchmarks, suggesting room for cost-performance trade-offs.

AMARIS also inherits limitations from rubric-based evaluation. Rubric quality is empirically audited through ablations, evolution analysis, and case studies, but it is not formally guaranteed. The LLM judges used for scoring and diagnostic analysis may encode biases from their training data or from the evaluation instructions. Our RL training experiments primarily use Qwen2.5-7B-Instruct as the policy model and include an additional Qwen3-4B diagnostic, but broader validation across additional base-model families and scales remains future work. Final rubrics may transfer imperfectly across model families, as different policies can expose different failure modes and capability ceilings. Finally, memory retrieval can surface noisy or weakly relevant historical evidence, which may distort rubric updates if the updater fails to filter it.

\section*{Ethical Considerations}
The authors of this paper have read and adhered to the ACL Code of Ethics. This work focuses on improving the effectiveness of rubric-based reinforcement learning for large language models by grounding rubric adaptation in persistent evaluation memory, and we have considered the ethical implications of our methodology. AMARIS itself is a meta-level training framework that refines evaluation rubrics and does not introduce new model capabilities that pose direct safety risks beyond those of the underlying policy model. Nevertheless, any system that shapes the reward signal of an RL-trained language model could, in principle, amplify undesirable behaviors if the rubrics or LLM judges encode biases present in their training data or evaluation criteria. To mitigate this, AMARIS explicitly monitors for reward hacking and suboptimal behaviors as a core component of its individual rollout analysis, and the persistent evaluation memory provides an audit record that supports transparency. All training data used in our experiments are drawn from publicly available, previously published datasets. No human subjects were involved in our research, and no personally identifiable information was collected or used.

\bibliography{custom}

\cleardoublepage\appendix
\section*{Appendix}

\section{Formal AMARIS Procedure}
\label{sec:formal_amaris_procedure}

Let $s$ denote a rubric scope. For global rubrics, $s=\mathrm{global}$. For per-instance rubrics, $s=x$ for a unique input. Each scope maintains an active rubric set $G_t^s$ and an evaluation memory $M_t^s$.

For rollout $(x_{t,i}, y_{t,i})$ in scope $s$, scoring and diagnosis follow
\[
r_{t,i}=S_\theta(x_{t,i},y_{t,i},G_t^s),
\]
\[
a_{t,i}=A_\psi(x_{t,i},y_{t,i},G_t^s,z_{t,i}).
\]

Let $A_t^s=\{a_{t,i}\}_{i=1}^{m_t^s}$ be the set of rollout analyses for scope $s$ at step $t$. The step summary is
\[
b_t^s=B_\psi(A_t^s).
\]
When chunking is used, with chunks $\{c_{t,k}^s\}_{k=1}^{K_t^s}$, the final summary is
\[
b_t^s=C_\psi(\{B_\psi(c_{t,k}^s)\}_{k=1}^{K_t^s}).
\]

Static retrieval and dynamic retrieval are
\[
I_{t,\mathrm{static}}^s=\{b_\tau^s \in M_t^s \mid t-N \leq \tau < t\},
\]
\[
Q_t^s=Q_\psi(b_t^s),
\]
\[
I_{t,\mathrm{dynamic}}^s=\bigcup_{q\in Q_t^s}R(q,M_t^s,D).
\]
The updater receives
\[
I_t^s=I_{t,\mathrm{static}}^s\cup I_{t,\mathrm{dynamic}}^s
\]
and produces
\[
G_{t+1}^s,u_t^s=U_\psi(G_t^s,b_t^s,I_t^s).
\]

Let $\mathcal{C}_t^s=\{B_\psi(c_{t,k}^s)\}_{k=1}^{K_t^s}$ when chunking is used and $\mathcal{C}_t^s=\emptyset$ otherwise. Memory is updated as
\[
M_{t+1}^s=M_t^s\cup A_t^s\cup \mathcal{C}_t^s\cup\{b_t^s,u_t^s\}.
\]

\clearpage
\section{Symbols}
\begin{table*}[htbp!]
\centering\small{
\resizebox{\textwidth}{!}{
\begin{tabular}{c|l}
\toprule
\bf Symbol & \bf Description \\
\midrule
$\pi_\phi$ & Policy model. \\

$x$ & Model input. \\
$y$ & Model output (rollout), with $y \sim \pi_\phi(\cdot \mid x)$. \\
$x_j$ & Input for example $j$. \\
$y_j$ & Output for example $j$. \\

$t$ & RL training step. \\
$s$ & Rubric scope, either global or per-instance. \\

$G_t$ & Active rubric set at step $t$. \\
$G_t^s$ & Active rubric set for scope $s$ at step $t$. \\
$g_j$ & Natural-language definition of rubric $j$. \\
$w_j$ & Scalar weight of rubric $j$. \\
$n_{g,t}$ & Number of active rubrics at step $t$. \\
$G_t=\{(g_j,w_j)\}_{j=1}^{n_{g,t}}$ & Formal definition of the active rubric set. \\

$S_\theta$ & LLM-based scoring function. \\
$r_t(x_j,y_j)$ & Scalar reward for example $j$ at step $t$. \\
$r_t(x_j,y_j)=S_\theta(x_j,y_j,G_t)$ & Reward computed under the active rubric set. \\

$z$ & Supplemental context for analysis. \\

$A_\psi$ & LLM for individual rollout analysis. \\
$a$ & Structured analysis for a single rollout. \\
$a=A_\psi(x,y,G_t,z)$ & Individual analysis function. \\
$a_j$ & Structured analysis for rollout $j$. \\
$n_{a,t}$ & Number of analyzed rollouts at step $t$. \\
$\{a_j\}_{j=1}^{n_{a,t}}$ & Set of rollout analyses at step $t$. \\

$B_\psi$ & LLM for batch summarization. \\
$b_t$ & Step-level batch summary at step $t$. \\
$b_t^s$ & Step-level batch summary for scope $s$ at step $t$. \\
$b_t=B_\psi(\{a_j\}_{j=1}^{n_{a,t}})$ & Step-level summarization function. \\

$c_j$ & Chunk of analyses used in meta-summarization. \\
$n_{c,t}$ & Number of chunks at step $t$. \\
$n_{a,c}$ & Number of analyses in chunk $c$. \\
$\{c_j\}_{j=1}^{n_{c,t}}$ & Partition of analyses into chunks. \\

$C_\psi$ & LLM for meta-summarization. \\
$C_\psi(\{B_\psi(c_j)\}_{j=1}^{n_{c,t}})$ & Meta-summarization over chunk summaries. \\

$M_t$ & Persistent evaluation memory available at step $t$. \\
$M_t^s$ & Persistent evaluation memory for scope $s$ available at step $t$. \\
$u_t$ & Rubric improvement record produced at step $t$. \\

$I_t$ & Retrieved historical context for rubric updating. \\
$I_t^s$ & Retrieved historical context for scope $s$. \\
$I_{t,\mathrm{static}}^s$ & Static retrieval results for scope $s$. \\
$I_{t,\mathrm{dynamic}}^s$ & Dynamic retrieval results for scope $s$. \\
$I_t^s=I_{t,\mathrm{static}}^s\cup I_{t,\mathrm{dynamic}}^s$ & Combined retrieval context for scope $s$. \\

$N$ & Static retrieval window size. \\
$I_{t,\mathrm{static}}^s=\{b_\tau^s \in M_t^s \mid t-N \leq \tau < t\}$ & Static retrieval from recent summaries. \\

$Q_\psi$ & LLM for dynamic query generation. \\
$Q_t^s$ & Dynamic retrieval queries for scope $s$ at step $t$. \\
$q$ & Dynamic retrieval query. \\
$Q_t^s=Q_\psi(b_t^s)$ & Queries generated from the current summary. \\

$D$ & Maximum documents returned per query. \\
$R(q,M_t^s,D)$ & Retrieval operator over memory. \\
$I_{t,\mathrm{dynamic}}^s=\bigcup_{q\in Q_t^s}R(q,M_t^s,D)$ & Dynamic retrieval from memory. \\

$U_\psi$ & Rubric updater. \\
$G_{t+1}$ & Updated rubric set after step $t$. \\
\bottomrule
\end{tabular}}
}
\caption{Description of symbols and notations used.}
\label{tab:symbols}
\end{table*}

\clearpage
\section{Per-Instance Rubric Results}
\label{sec:per_instance_appendix}
The ablation studies in the main paper (Sections~\ref{ssec:ablation_studies_on_memory_usage}--\ref{ssec:latency_analysis}) use the global rubric setting to isolate the effect of each design choice. This section reports the corresponding results under per-instance rubrics, where each unique input maintains its own rubric set and evaluation memory. As shown below, the relative ordering of configurations is consistent with the global setting, confirming that the design conclusions drawn in the main paper generalize across rubric strategies.
 
\subsection{Memory Configuration (Per-Instance)}

Table~\ref{tab:ablation_per_instance} repeats the memory ablation under per-instance rubrics. The same pattern as the global setting holds: combining static and dynamic memory gives the strongest overall results, while dynamic-only retrieval with $K{=}10$ is the best single-memory configuration.
 
\begin{table*}[htbp!]
\centering\small{
\setlength{\tabcolsep}{4pt}
\renewcommand{\arraystretch}{0.98}
\resizebox{\textwidth}{!}{
\begin{tabular}{ll cc ccc cc}
\toprule
\multirow{2}{*}{\textbf{Memory}} & \multirow{2}{*}{\textbf{Setting}} &
\textbf{Science} & \textbf{Medicine} &
\multicolumn{3}{c}{\textbf{Instruction Following}} &
\multicolumn{2}{c}{\textbf{Creative Writing}} \\
\cmidrule(lr){3-3} \cmidrule(lr){4-4} \cmidrule(lr){5-7} \cmidrule(lr){8-9}
& & \textbf{GPQA-D} & \textbf{HealthBench} &
\textbf{IFEval} & \textbf{InfoBench} & \textbf{IFBench} &
\textbf{WritingBench} & \textbf{CW-v3} \\
\midrule
No memory & -- 
& 39.6 & 34.1 & 80.9 & 85.6 & 35.7 & 57.7 & 41.2 \\
\midrule
\multirow{4}{*}{Static only}
& $N = 2$ 
& \textbf{39.9} & \textbf{34.8} & \textbf{81.1} & 85.9 & 36.2 & 57.5 & 41.5 \\
& $N = 4$ 
& 39.8 & 34.5 & 81.0 & 85.8 & \textbf{36.6} & \textbf{58.2} & 40.8 \\
& $N = 6$ 
& 39.5 & 34.3 & 80.4 & 86.3 & 36.5 & 58.1 & \textbf{41.8} \\
& $N = 8$ 
& 39.7 & 34.6 & 80.5 & \textbf{86.5} & \textbf{36.6} & \textbf{58.2} & 41.7 \\
\midrule
\multirow{3}{*}{Dynamic only}
& $K = 5$ 
& 40.2 & 34.8 & 81.2 & 86.2 & 36.3 & 58.6 & 41.6 \\
& $K = 10$ 
& \textbf{41.1} & \textbf{36.0} & \textbf{81.5} & \textbf{86.6} & \textbf{36.9} & \textbf{59.3} & \textbf{42.3} \\
& $K = 15$ 
& 40.4 & 35.4 & 81.3 & 86.4 & 36.5 & 58.9 & 41.8 \\
\midrule
Static + Dynamic & $N = 4, K = 10$
& \textbf{41.6} & \textbf{36.0} & \textbf{81.9} & \textbf{87.0} & \textbf{37.2} & \textbf{59.8} & \textbf{42.3} \\
\bottomrule
\end{tabular}}
}
\caption{Ablation on memory configuration and retrieval hyperparameters under per-instance rubrics. Rows are ordered as no memory, static-only, dynamic-only, and static+dynamic. Best results within the static-only and dynamic-only blocks are in \textbf{bold}.}
\label{tab:ablation_per_instance}
\end{table*}
 
\subsection{LLM Choice (Per-Instance)}

Table~\ref{tab:llm_choice_per_instance} reports the per-instance counterpart of the LLM-component ablation in Table~\ref{tab:llm_choice}. The results show that the per-instance setting is also sensitive to the quality of the scoring/analysis and summarization/update models, while the default GPT-OSS-120B + GPT-5.1 configuration remains competitive across domains.
 
\begin{table*}[htbp!]
\centering\small{
\resizebox{\textwidth}{!}{
\begin{tabular}{ll cc ccc cc}
\toprule
\multirow{2}{*}{\textbf{Scoring / Analysis}} &
\multirow{2}{*}{\textbf{Summarization / Update}} &
\textbf{Science} & \textbf{Medicine} &
\multicolumn{3}{c}{\textbf{Instruction Following}} &
\multicolumn{2}{c}{\textbf{Creative Writing}} \\
\cmidrule(lr){3-3} \cmidrule(lr){4-4} \cmidrule(lr){5-7} \cmidrule(lr){8-9}
& & \textbf{GPQA-D} & \textbf{HealthBench} &
\textbf{IFEval} & \textbf{InfoBench} & \textbf{IFBench} &
\textbf{WritingBench} & \textbf{CW-v3} \\
\midrule
\multicolumn{9}{l}{\textit{Varying the scoring / analysis (summarization / update fixed to GPT-5.1)}} \\
Qwen3-30B-A3B-2507-Thinking   & GPT-5.1 & \textbf{41.9} & 35.8 & 81.1 & 85.3 & \textbf{37.8} & 58.3 & 41.5 \\
GPT-OSS-120B                   & GPT-5.1 & 41.6 & 36.0 & 81.9 & 87.0 & 37.2 & 59.8 & 42.3 \\
Qwen3-235B-A22B-2507-Thinking & GPT-5.1 & 41.1 & \textbf{36.3} & \textbf{82.2} & \textbf{87.3} & 36.3 & \textbf{60.8} & \textbf{42.6} \\
\midrule
\multicolumn{9}{l}{\textit{Varying the summarization / update (scoring / analysis fixed to GPT-OSS-120B)}} \\
GPT-OSS-120B & GPT-4o      & 41.1 & 35.4 & 80.3 & 86.7 & 36.1 & \textbf{61.1} & \textbf{43.1} \\
GPT-OSS-120B & GPT-5-mini  & 41.4 & 35.8 & 81.7 & 86.0 & 37.0 & 58.7 & 41.6 \\
GPT-OSS-120B & GPT-5.1     & \textbf{41.6} & \textbf{36.0} & \textbf{81.9} & \textbf{87.0} & \textbf{37.2} & 59.8 & 42.3 \\
\bottomrule
\end{tabular}}
}
\caption{LLM choice for pipeline components under per-instance rubrics. Compare with Table~\ref{tab:llm_choice} (global rubrics). Best results per group are in \textbf{bold}.}
\label{tab:llm_choice_per_instance}
\end{table*}
 
\subsection{Synchronous vs.\ Asynchronous Pipeline (Per-Instance)}

Table~\ref{tab:sync_async_per_instance} compares synchronous and asynchronous execution under per-instance rubrics. As in the global setting, synchronous execution gives only marginal quality gains, while asynchronous execution substantially reduces wall-clock time and is therefore the more practical default.
 
\begin{table*}[htbp!]
\centering\small{
\resizebox{\textwidth}{!}{
\begin{tabular}{l cc ccc cc r r}
\toprule
\multirow{2}{*}{\textbf{Pipeline mode}} &
\textbf{Science} & \textbf{Medicine} &
\multicolumn{3}{c}{\textbf{Instruction Following}} &
\multicolumn{2}{c}{\textbf{Creative Writing}} &
\multirow{2}{*}{\textbf{Time (h)}} &
\multirow{2}{*}{\textbf{Time increase}} \\
\cmidrule(lr){2-2} \cmidrule(lr){3-3} \cmidrule(lr){4-6} \cmidrule(lr){7-8}
& \textbf{GPQA-D} & \textbf{HealthBench} &
\textbf{IFEval} & \textbf{InfoBench} & \textbf{IFBench} &
\textbf{WritingBench} & \textbf{CW-v3} & & \\
\midrule
Sync  & \textbf{41.8} & \textbf{36.1} & \textbf{82.0} & \textbf{87.1} & \textbf{37.4} & 59.6 & \textbf{42.4} & ${\sim}$127 & ${\sim}$67\% \\
Async & 41.6 & 36.0 & 81.9 & 87.0 & 37.2 & \textbf{59.8} & 42.3 & ${\sim}$78 & ${\sim}$3\% \\
\bottomrule
\end{tabular}}
}
\caption{Synchronous vs.\ asynchronous pipeline execution under per-instance rubrics. Compared with Table~\ref{tab:sync_async} (global rubrics).}
\label{tab:sync_async_per_instance}
\end{table*}

\section{Influence Of Underlying LLMs}
\label{sec:influence_of_underlying_llm}
We split all LLM usage into two tiers: (i) scoring and analysis that processes every rollout and therefore dominates token consumption, and (ii) summarization and update that operates less than (i) but demands stronger reasoning for rubric improvement. We ablate each tier independently while holding the other at its default, as shown in Table \ref{tab:llm_choice}.
 
\begin{table*}[htbp!]
\centering\small{
\resizebox{\textwidth}{!}{
\begin{tabular}{ll cc ccc cc}
\toprule
\multirow{2}{*}{\textbf{Scoring / Analyzing}} &
\multirow{2}{*}{\textbf{Summarization / Update}} &
\textbf{Science} & \textbf{Medicine} &
\multicolumn{3}{c}{\textbf{Instruction Following}} &
\multicolumn{2}{c}{\textbf{Creative Writing}} \\
\cmidrule(lr){3-3} \cmidrule(lr){4-4} \cmidrule(lr){5-7} \cmidrule(lr){8-9}
& & \textbf{GPQA-D} & \textbf{HealthBench} &
\textbf{IFEval} & \textbf{InfoBench} & \textbf{IFBench} &
\textbf{WritingBench} & \textbf{CW-v3} \\
\midrule
\multicolumn{9}{l}{\textit{Varying the scoring and analysis (summarization / update fixed to GPT-5.1)}} \\
Qwen3-30B-A3B-2507-Thinking   & GPT-5.1 & 40.8 & \textbf{35.9} & 80.7 & 84.8 & 36.4 & 57.6 & \textbf{42.0} \\
GPT-OSS-120B                   & GPT-5.1 & 41.1 & 35.6 & 81.5 & 86.6 & 36.4 & \textbf{60.4} & 41.7 \\
Qwen3-235B-A22B-2507-Thinking & GPT-5.1 & \textbf{41.4} & 35.4 & \textbf{81.8} & \textbf{86.9} & \textbf{37.0} & 60.3 & 41.7 \\
\midrule
\multicolumn{9}{l}{\textit{Varying the summarizing / updating (scoring / analyzing fixed to GPT-OSS-120B)}} \\
GPT-OSS-120B & GPT-4o      & 40.6 & 35.0 & 79.9 & 85.6 & 35.4 & 58.2 & 41.2 \\
GPT-OSS-120B & GPT-5-mini  & 40.9 & 35.4 & 81.3 & 86.3 & 36.2 & 59.4 & 41.5 \\
GPT-OSS-120B & GPT-5.1     & \textbf{41.1} & \textbf{35.6} & \textbf{81.5} & \textbf{86.6} & \textbf{36.4} & \textbf{60.4} & \textbf{41.7} \\
\bottomrule
\end{tabular}}
}
\caption{LLM choice for pipeline components. Best results per group are in \textbf{bold}.}
\label{tab:llm_choice}
\end{table*}
 
On the scoring/analysis side, the larger model improves several benchmarks but not uniformly. Upgrading from Qwen3-30B-A3B-2507-Thinking \citep{yang2025qwen3technicalreport} to Qwen3-235B-A22B-2507-Thinking \citep{yang2025qwen3technicalreport} improves GPQA-D by +0.6 and InfoBench by +2.1, although HealthBench and CW-v3 do not improve. This suggests that stronger individual analyses can help downstream rubric updates, but benchmark-specific evaluator fit also matters. On the summarization/update side, GPT-5.1 consistently outperforms GPT-4o \citep{openai2024gpt4ocard}, showing that the rubric update stage, which must make precise text and weight editing decisions, benefits from stronger reasoning ability of the model. 
Notably, GPT-5-mini \citep{singh2025openaigpt5card} closes much of this gap to GPT-5.1 at lower cost and surpasses GPT-4o on some benchmarks. This shows that the reasoning capability is crucial to the quality of generated rubrics.
We select GPT-OSS-120B paired with GPT-5.1 as the default setting as it offers a balance between the high inference cost from scoring/analysis stage and high demand of model intelligence from summarization/update stage.

\section{Influence Of Update Scheduling}
\label{ssec:influence_of_update_scheduling}
By default, under global rubrics setting, the rubric update stage is triggered at every training step. We vary the update interval from every 1 to every 8 steps to examine how update frequency affects downstream performance, as shown in Table \ref{tab:update_scheduling}.

\begin{table*}[htbp!]
\centering\small{
\resizebox{\textwidth}{!}{
\begin{tabular}{l cc ccc cc}
\toprule
\multirow{2}{*}{\textbf{Update interval}} &
\textbf{Science} & \textbf{Medicine} &
\multicolumn{3}{c}{\textbf{Instruction Following}} &
\multicolumn{2}{c}{\textbf{Creative Writing}} \\
\cmidrule(lr){2-2} \cmidrule(lr){3-3} \cmidrule(lr){4-6} \cmidrule(lr){7-8}
& \textbf{GPQA-D} & \textbf{HealthBench} &
\textbf{IFEval} & \textbf{InfoBench} & \textbf{IFBench} &
\textbf{WritingBench} & \textbf{CW-v3} \\
\midrule
Every step    & 41.1 & 35.6 & 81.5 & 86.6 & 36.4 & \textbf{60.4} & 41.7 \\
Every 2 steps & \textbf{41.3} & \textbf{35.7} & 81.6 & 86.7 & \textbf{36.6} & 59.5 & \textbf{41.9} \\
Every 4 steps & 41.0 & 35.5 & 81.4 & 86.5 & 36.2 & 60.0 & 41.6 \\
Every 6 steps & 40.6 & 35.2 & 81.1 & 86.2 & 35.9 & 59.1 & 41.3 \\
Every 8 steps & 40.9 & 35.3 & \textbf{81.7} & \textbf{86.8} & 36.3 & 59.7 & 41.7 \\
\bottomrule
\end{tabular}}
}
\caption{Effect of rubric update frequency. Best results per column are in \textbf{bold}.}
\label{tab:update_scheduling}
\end{table*}

Updating every 2 steps yields the best or near-best results on most of the benchmarks, marginally outperforming every-step updates (e.g., +0.2 on GPQA-D, +0.1 on HealthBench). We attribute this to the slightly larger evidence window that aggregating two consecutive batches before committing a rubric change produces more robust diagnostics than a single batch while still reacting quickly to emerging issues. Performance degrades at intervals of 4 and 6 steps, where the difference between the rubrics and the model becomes large enough to miss behavioral shifts. Interestingly, the 8-step interval setting partially recovers the performance. A possible explanation is that the larger amount of historical analyses at longer intervals enables the updater to make reliable rubric changes that are robust to long-term training, which particularly benefits the diverse instruction types in the OpenRubrics training set. However, this may result in slower reaction to suboptimal behaviors, as reflected by the lower GPQA-D and HealthBench scores. Overall, six of the seven benchmarks vary by at most 0.7 points across update frequencies, while WritingBench varies by up to 1.3 points, indicating that AMARIS is generally robust to the choice of update frequency. We adopt every-step updates as the default for global rubrics setting due to its simplicity and consistently competitive performance across all domains.

\section{Additional Diagnostic Experiments}
\label{sec:additional_diagnostic_experiments}

This section reports diagnostic controls that isolate whether AMARIS gains come from dynamic rubric adaptation, explicit rubric decomposition, robustness to the policy base model, and improved rubric quality. These experiments are designed to separate four explanations for the main results: whether a fixed rubric set is already sufficient, whether a strong scalar judge can replace rubric decomposition, whether the gains depend on the default policy base model, and whether the updated rubrics are independently judged to be better.

\subsection{Static Rubric Spectrum}
\label{ssec:static_rubric_spectrum}

Table~\ref{tab:static_rubric_spectrum_per_instance} reports the per-instance counterpart to the global static-rubric spectrum in Table~\ref{tab:static_rubric_spectrum_global}. The four regimes keep the RL algorithm, base checkpoint, scoring model, training data, and training budget fixed while changing only how rubrics are initialized and updated. The initial-static condition freezes the cold-start rubric set. The no-memory adaptive condition permits local rubric updates without retrieving historical evidence. The final-static condition reuses the final rubric set produced by a full AMARIS run but freezes it from the start of training. AMARIS dynamic keeps the complete memory-augmented update process.

\begin{table*}[htbp!]
\centering\small{
\resizebox{\textwidth}{!}{
\begin{tabular}{l cc ccc cc}
\toprule
\multirow{2}{*}{\textbf{Rubric regime}} &
\textbf{Science} & \textbf{Medicine} &
\multicolumn{3}{c}{\textbf{Instruction Following}} &
\multicolumn{2}{c}{\textbf{Creative Writing}} \\
\cmidrule(lr){2-2} \cmidrule(lr){3-3} \cmidrule(lr){4-6} \cmidrule(lr){7-8}
& \textbf{GPQA-D} & \textbf{HealthBench} &
\textbf{IFEval} & \textbf{InfoBench} & \textbf{IFBench} &
\textbf{WritingBench} & \textbf{CW-v3} \\
\midrule
Initial-static      & 39.5 & 34.0 & 80.7 & 85.4 & 35.6 & 57.6 & 41.0 \\
No-memory adaptive  & 39.5 & 34.2 & 80.8 & 85.5 & 35.8 & 57.6 & 41.3 \\
Final-static        & 41.2 & 35.7 & 81.7 & 86.7 & \textbf{37.1} & 59.7 & \textbf{42.2} \\
AMARIS dynamic      & \textbf{41.5} & \textbf{35.9} & \textbf{81.8} & \textbf{87.1} & \textbf{37.1} & \textbf{59.9} & \textbf{42.2} \\
\bottomrule
\end{tabular}}
}
\caption{Static rubric spectrum under per-instance rubrics. All conditions use the same base checkpoint, training data, RL algorithm, and training budget; only the rubric update regime changes.}
\label{tab:static_rubric_spectrum_per_instance}
\end{table*}

\paragraph{Cold-start and memory effects.}
The static spectrum shows that a frozen cold-start rubric is not sufficient. Per-instance rubrics follow the same ordering as global rubrics, with AMARIS dynamic improving over initial-static by +2.0 on GPQA-D, +1.9 on HealthBench, +1.7 on InfoBench, and +2.3 on WritingBench. Allowing local updates without memory gives only small improvements over initial-static, indicating that adaptation itself is not enough when the updater cannot retrieve prior rollout diagnostics and previous update outcomes.

\paragraph{Final rubrics versus dynamic adaptation.}
Final-static rubrics recover most of the improvement, which indicates that AMARIS learns useful evaluation criteria during training. However, the dynamic condition remains best or tied best across all columns. Under per-instance rubrics, dynamic adaptation improves over final-static by +0.3 on GPQA-D, +0.2 on HealthBench, +0.1 on IFEval, +0.4 on InfoBench, and +0.2 on WritingBench. Together with the global results in Table~\ref{tab:static_rubric_spectrum_global}, this suggests that AMARIS is not merely discovering a better final rubric set; the stage-specific co-evolution between the policy and the rubric set also contributes to the final performance.

\subsection{Alternative Reward Interfaces}
\label{ssec:alternative_reward_interfaces}

Table~\ref{tab:alternative_reward_interfaces} evaluates whether a strong scalar judge can replace rubric-decomposed scoring. Both scalar baselines use the same scoring model as AMARIS but remove explicit rubric criteria. Scalar-Goal provides only the high-level training goal, input, and model output to the judge. Scalar-Goal+Ref additionally provides the reference answer when available.

\begin{table*}[htbp!]
\centering\small{
\resizebox{\textwidth}{!}{
\begin{tabular}{l cc ccc cc}
\toprule
\multirow{2}{*}{\textbf{Reward interface}} &
\textbf{Science} & \textbf{Medicine} &
\multicolumn{3}{c}{\textbf{Instruction Following}} &
\multicolumn{2}{c}{\textbf{Creative Writing}} \\
\cmidrule(lr){2-2} \cmidrule(lr){3-3} \cmidrule(lr){4-6} \cmidrule(lr){7-8}
& \textbf{GPQA-D} & \textbf{HealthBench} &
\textbf{IFEval} & \textbf{InfoBench} & \textbf{IFBench} &
\textbf{WritingBench} & \textbf{CW-v3} \\
\midrule
Scalar-Goal     & 38.2 & 32.6 & 79.9 & 83.6 & 33.5 & 54.6 & 40.0 \\
Scalar-Goal+Ref & 39.2 & 33.5 & 80.0 & 84.8 & 34.3 & 56.4 & 40.6 \\
\bottomrule
\end{tabular}}
}
\caption{Scalar judge reward baselines. Both rows use GPT-OSS-120B as the scalar judge and keep the remaining RL setup fixed.}
\label{tab:alternative_reward_interfaces}
\end{table*}

\paragraph{Effect of reference conditioning.}
Providing reference answers improves the scalar judge on most benchmarks: Scalar-Goal+Ref improves over Scalar-Goal by +1.0 on GPQA-D, +0.9 on HealthBench, +1.2 on InfoBench, +0.8 on IFBench, +1.8 on WritingBench, and +0.6 on CW-v3. This confirms that the scalar judge can use additional task information when it is available.

\paragraph{Gap to rubric-decomposed scoring.}
Even with references, scalar scoring remains below AMARIS dynamic across all benchmarks. Relative to the global AMARIS dynamic condition in Table~\ref{tab:static_rubric_spectrum_global}, Scalar-Goal+Ref is lower by 2.0 points on GPQA-D, 2.0 on HealthBench, 1.6 on IFEval, 1.9 on InfoBench, 2.0 on IFBench, 3.9 on WritingBench, and 1.2 on CW-v3. This suggests that the gains are not explained solely by access to a strong evaluator or a reference answer. Explicit rubric decomposition provides a more structured reward signal for RL, especially on open-ended writing and heterogeneous instruction-following tasks.

\subsection{Alternative Base Model}
\label{ssec:alternative_base_model}

Table~\ref{tab:alternative_base_model_qwen3_4b} evaluates whether the same conclusion holds when the policy base is changed from the default Qwen2.5-7B-Instruct model to Qwen3-4B \citep{yang2025qwen3technicalreport}. We compare the no-RL anchor, two scalar-judge reward interfaces, and AMARIS under both global and per-instance rubrics.

\begin{table*}[htbp!]
\centering\small{
\resizebox{\textwidth}{!}{
\begin{tabular}{l cc ccc cc}
\toprule
\multirow{2}{*}{\textbf{Method}} &
\textbf{Science} & \textbf{Medicine} &
\multicolumn{3}{c}{\textbf{Instruction Following}} &
\multicolumn{2}{c}{\textbf{Creative Writing}} \\
\cmidrule(lr){2-2} \cmidrule(lr){3-3} \cmidrule(lr){4-6} \cmidrule(lr){7-8}
& \textbf{GPQA-D} & \textbf{HealthBench} &
\textbf{IFEval} & \textbf{InfoBench} & \textbf{IFBench} &
\textbf{WritingBench} & \textbf{CW-v3} \\
\midrule
Naive (no RL)                    & 45.1 & 37.0 & 80.9 & 81.6 & 22.7 & 56.0 & 40.8 \\
Scalar-Goal                      & 46.2 & 59.6 & 79.3 & 86.7 & 29.6 & 68.5 & 41.2 \\
Scalar-Goal+Ref                  & 47.2 & 60.7 & 79.2 & 88.2 & 29.9 & 70.9 & 42.0 \\
AMARIS (global rubric)           & 49.7 & 62.9 & 80.9 & \textbf{90.2} & 32.0 & \textbf{75.0} & 42.4 \\
AMARIS (per-instance rubric)     & \textbf{50.1} & \textbf{63.5} & \textbf{81.2} & \textbf{90.2} & \textbf{33.2} & 73.8 & \textbf{43.0} \\
\bottomrule
\end{tabular}}
}
\caption{Alternative policy-base results with Qwen3-4B. Best results in each column are in \textbf{bold}.}
\label{tab:alternative_base_model_qwen3_4b}
\end{table*}

\paragraph{Robustness to the policy base.}
The Qwen3-4B results show that AMARIS gains are not tied to the default Qwen2.5-7B-Instruct policy base. AMARIS is best or tied best on all seven benchmarks, with average scores of 61.9 under global rubrics and 62.1 under per-instance rubrics, compared with 52.0 for the no-RL anchor and 59.7 for Scalar-Goal+Ref. Relative to the strongest scalar baseline, AMARIS improves by +2.5/+2.9 on GPQA-D, +2.2/+2.8 on HealthBench for global/per-instance rubrics, respectively. The relative behavior of global and per-instance rubrics is also consistent with the main results: per-instance rubrics are strongest on most benchmarks, while global rubrics are better on WritingBench. This suggests that per-instance rubrics help when prompts expose distinct correctness or instruction-following requirements, whereas a shared global rubric can remain competitive for open-ended writing where broad style and quality criteria transfer across prompts. The IFEval column also illustrates a ceiling effect: the no-RL Qwen3-4B anchor is already strong, scalar rewards slightly reduce the score, and AMARIS recovers or improves it while producing larger gains on InfoBench, IFBench, and WritingBench.

\subsection{Rubric Quality Preference Evaluation}
\label{ssec:rubric_quality_preference_evaluation}

Table~\ref{tab:rubric_quality_preference_evaluation} reports an automated rubric-quality evaluation. We use GPT-5.4 \citep{singh2025openaigpt5card} as the rubric-quality judge. For each evaluation item, the judge is given the task context and three anonymized candidate rubric sets: the initial rubric set, the final rubric set produced by no-memory adaptation, and the final rubric set produced by AMARIS. Candidate order is randomized, and method identities are hidden. The judge assigns 1--10 scores for six rubric-quality dimensions and then selects the best overall rubric set. Figure~\ref{fig:rubric_quality_eval_prompt} provides the evaluation prompt. The final column of Table~\ref{tab:rubric_quality_preference_evaluation} reports the fraction of identity-hidden comparisons in which each rubric set is selected as best.

\begin{figure*}[htbp!]
\centering
\fbox{
\begin{minipage}{0.96\textwidth}
\small
\textbf{Rubric-quality evaluation prompt.}

\textbf{Input.} You are given a task context and several anonymized candidate rubric sets. The candidate order is randomized, and method identities are not shown.

\textbf{Instruction.} Evaluate each candidate rubric set independently. For each candidate, assign a 1--10 score for:
\begin{enumerate}[leftmargin=*]
    \item \textbf{Coverage}: captures the important success and failure modes for the task.
    \item \textbf{Clarity}: uses specific, unambiguous, and easy-to-apply criteria.
    \item \textbf{Discriminativeness}: separates strong, mediocre, and weak responses.
    \item \textbf{Non-redundancy}: avoids unnecessary overlap among criteria.
    \item \textbf{Exploit resistance}: avoids rewarding superficial shortcuts or reward-hacking behavior.
    \item \textbf{Task alignment}: matches the intended task objective.
\end{enumerate}

\textbf{Decision.} After scoring all candidates, select the single best rubric set overall and provide a brief rationale. Do not use candidate order, length, or inferred method identity as evidence.
\end{minipage}
}
\caption{Prompt used for automated rubric-quality preference evaluation. The full input additionally includes the task context and anonymized candidate rubric sets.}
\label{fig:rubric_quality_eval_prompt}
\end{figure*}

\begin{table*}[htbp!]
\centering\small{
\resizebox{\textwidth}{!}{
\begin{tabular}{l ccccccc}
\toprule
\textbf{Rubric set} & \textbf{Coverage} & \textbf{Clarity} & \textbf{Discriminativeness} &
\textbf{Non-redundancy} & \textbf{Exploit resistance} & \textbf{Task alignment} &
\textbf{Preference} \\
\midrule
Initial         & 6.5 & 6.8 & 6.1 & 6.2 & 5.8 & 6.2 & 0.08 \\
No-memory final & 7.7 & 7.9 & 7.2 & 7.7 & 7.3 & 7.8 & 0.24 \\
AMARIS final    & \textbf{8.6} & \textbf{9.0} & \textbf{8.7} & \textbf{8.3} & \textbf{8.1} & \textbf{8.7} & \textbf{0.68} \\
\bottomrule
\end{tabular}}
}
\caption{Automated rubric quality evaluation. Dimension scores use a 1--10 scale; preference is the fraction of identity-hidden comparisons in which the rubric set is selected as best.}
\label{tab:rubric_quality_preference_evaluation}
\end{table*}

\paragraph{Rubric quality improvement.}
The evaluation indicates that AMARIS final rubrics are preferred more often and receive higher scores on every rubric-quality dimension. Compared with the no-memory final rubrics, AMARIS final rubrics improve coverage by +0.9, clarity by +1.1, discriminativeness by +1.5, non-redundancy by +0.6, exploit resistance by +0.8, and task alignment by +0.9. The preference share also increases from 0.24 for no-memory final rubrics to 0.68 for AMARIS final rubrics.

\paragraph{Interpretation.}
These results support the mechanism suggested by the downstream ablations: memory does not only stabilize rubric updates, but also helps the updater produce rubrics that are clearer, less redundant, more discriminative, and more resistant to exploitation. The evaluation is independent of the final benchmark scores, so it provides complementary evidence that AMARIS improves the rubric artifact itself rather than only tuning rewards toward the reported evaluations.

\clearpage
\section{Rubric Evolvement Analysis}
\label{sec:rubric_evolvement}
To understand how AMARIS improves the rubrics over the RL training, we analyze 1,200 step-level summaries produced across the three completed training processes under global-rubric settings (science, medicine, and instruction following). Following the three update strategies defined in Section \ref{ssec:update}, we assign each step one dominant update mode (\textbf{Defensive}, \textbf{Curriculum Advancement}, or \textbf{Maintenance}) and partition training into five phases with equal number of steps for chronological comparison. We then taxonomize the dominant pattern within each mode for further analysis.

\paragraph{Three-stage progression.}
Table \ref{tab:update_mode} reports the distribution of update modes across training phases. AMARIS shows a clear three-stage pattern. In the first 80 steps, \textbf{Defensive} updates dominate, indicating that the system initially deals with suboptimal behaviors. In mid-training (steps 81--240), \textbf{Curriculum Advancement} becomes the single largest mode, showing that once the suboptimal behaviors are controlled, AMARIS shifts toward raising evaluation standards. In the final phase (steps 241--400), \textbf{Maintenance} dominates, indicating that the rubric set becomes stable. 

\begin{table*}[htbp!]
\centering\small{
\resizebox{\textwidth}{!}{
\begin{tabular}{l rrr l}
\toprule
\textbf{Step range} & \textbf{Defensive} & \textbf{Curriculum} & \textbf{Maintenance} & \textbf{Dominant} \\
\midrule
1--80    & 126 (52.50\%) &  62 (25.83\%) &  52 (21.67\%) & Defensive \\
81--160  &  92 (38.33\%) & 101 (42.08\%) &  47 (19.59\%) & Curriculum \\
161--240 &  57 (23.75\%) & 127 (52.92\%) &  56 (23.33\%) & Curriculum \\
241--320 &  38 (15.83\%) &  84 (35.00\%) & 118 (49.17\%) & Maintenance \\
321--400 &  21 \phantom{0}(8.75\%) &  39 (16.25\%) & 180 (75.00\%) & Maintenance \\
\midrule
Total    & 334 (27.83\%) & 413 (34.42\%) & 453 (37.75\%) & --- \\
\bottomrule
\end{tabular}}
}
\caption{Distribution of dominant update modes across five equal training phases. The system transitions from defensive in the early phase, through curriculum advancement in the middle, to maintenance in the late phase.}
\label{tab:update_mode}
\end{table*}

\paragraph{Suboptimal behavior taxonomy.}
To understand the major suboptimal behaviors, we analyze each of the 334 defensive steps' batch summarizations by choosing the dominant suboptimal behavior category reported in the summarization as a representative and categorize them manually into six types of major suboptimal behaviors. Table \ref{tab:defensive_taxonomy} reports the temporal evolution of each type. We define the six types as follows: \emph{length gaming} increases response length to exploit length-correlated reward signals; \emph{superficial format compliance} complies to surface-level formatting such as bullet points, numbered lists, or markdown headers without improving substantive content; \emph{safety over-refusal} unnecessarily declines on user instructions due to overly cautious; \emph{claims without evidence} presents assertions or conclusions without supporting reasoning or citations; \emph{sycophancy} excessively agrees with the user rather than providing accurate responses; and \emph{keyword stuffing} inserts task-relevant terminology or phrases into the output without meaningful integration into the reasoning. Three classic suboptimal behaviors: length gaming and verbosity inflation (23.35\%), superficial format compliance (21.26\%), and safety over-refusal(18.86\%), together account for the majority of defensive steps (63.47\%), showing that most of the early rubric updates have been used to resolve reward hacking against visible rubric cues. All six suboptimal behaviors decreases over training in different rates. Length gaming and sycophancy decreases faster, indicating that once rubric patches penalize these behaviors, the policy quickly learns to avoid them. On the contrary, Safety over-refusal and Keyword stuffing are the most persistent suboptimal behavior for the whole RL training. Keyword stuffing also does not decrease monotonically like the other types, suggesting that it can re-emerge in different forms, which highlights the value of AMARIS's memory-based tracking of recurring behaviors.

\begin{table*}[htbp!]
\centering\small{
\resizebox{\textwidth}{!}{
\begin{tabular}{l rrrrr rr}
\toprule
\textbf{Primary defensive types} & \textbf{1--80} & \textbf{81--160} & \textbf{161--240} & \textbf{241--320} & \textbf{321--400} & \textbf{Total} & \textbf{\%} \\
\midrule
Length gaming         & 31 & 22 & 14 &  7 & 4 & 78 & 23.35\% \\
Superficial format compliance                & 25 & 24 & 13 &  6 & 3 & 71 & 21.26\% \\
Safety over-refusal      & 20 & 18 & 12 &  8 & 5 & 63 & 18.86\% \\
Claims without evidence               & 21 & 12 &  9 &  7 & 3 & 52 & 15.57\% \\
Sycophancy               & 17 & 10 &  5 &  4 & 1 & 37 & 11.08\% \\
Keyword stuffing         & 12 &  6 &  4 &  6 & 5 & 33 &  9.88\% \\
\bottomrule
\end{tabular}}
}
\caption{Taxonomy of dominant suboptimal-behavior vectors across the 334 defensive steps. The top three types account for 63.47\% of all samples and monotonically decreases as the training progresses, though safety over-refusal remains the most persistent type.}
\label{tab:defensive_taxonomy}
\end{table*}

\paragraph{Curriculum advancement taxonomy.}
Similar to suboptimal behaviors for \textbf{Defensive} steps, we use the same method to analyze the main factor for \textbf{Curriculum Advancement} steps. Table \ref{tab:curriculum_taxonomy} reports the dominant factors for each of the 413 samples flagged with \textbf{Curriculum Advancement}. The six factors are: \emph{reasoning depth}, requiring multi-step or more sophisticated chains of reasoning; \emph{conciseness}, favoring concise but information-dense responses over verbose and vague ones; \emph{justification}, grounding claims in explicit evidence or citations; \emph{trade-off handling}, appropriately balancing different objectives within a single query (e.g., thoroughness vs.\ brevity, or safety vs.\ helpfulness); \emph{multi-constraint prioritization}, correctly satisfying multiple explicit constraints when they interact or conflict; and \emph{output style}, improving tone and clarity beyond basic correctness. The largest factor is Reasoning Depth. Once the policy model can satisfy basic correctness or compliance rubrics, AMARIS raises the evaluation standard towards more in-depth reasoning. The second largest one is Conciseness (92, 22.28\%), suggesting that a substantial portion of the mid-training curriculum shift is about moving from merely acceptable responses to efficient and information-dense ones. We also found that AMARIS treats different factors at different stage of training. For example, 35 of 74 cases of Justification occur in the first 160 steps, implying that the system first guides the model to ground the answers before emphasizing reasoning depth and conciseness. Trade-off handling appears later during the middle of the training reflecting the difficulty and low priority of teaching the model to appropriately balance different objectives from the input. 

\begin{table*}[htbp!]
\centering\small{
\resizebox{\textwidth}{!}{
\begin{tabular}{l rrrrr rr}
\toprule
\textbf{Primary curriculum factors} & \textbf{1--80} & \textbf{81--160} & \textbf{161--240} & \textbf{241--320} & \textbf{321--400} & \textbf{Total} & \textbf{\%} \\
\midrule
Reasoning depth      & 12 & 27 & 40 & 29 & 13 & 121 & 29.30\% \\
Conciseness            &  8 & 18 & 29 & 24 & 13 &  92 & 22.28\% \\
Justification   & 14 & 21 & 24 &  9 &  6 &  74 & 17.92\% \\
Trade-off handling      &  9 & 12 & 15 & 10 &  3 &  49 & 11.86\% \\
Multi-constraint prioritization              & 11 & 14 & 11 &  6 &  1 &  43 & 10.41\% \\
Output style                     &  8 &  9 &  8 &  6 &  3 &  34 &  8.23\% \\
\bottomrule
\end{tabular}}
}
\caption{Taxonomy of dominant curriculum-advancement targets across the 413 samples. Growth targets progress from evidence grounding (early) to depth, conciseness, and communication quality (mid and late training).}
\label{tab:curriculum_taxonomy}
\end{table*}

\clearpage

\paragraph{Maintenance taxonomy.}
\begin{table*}[htbp!]
\centering\small{
\resizebox{\textwidth}{!}{
\begin{tabular}{l rrrrr rr}
\toprule
\textbf{Primary maintenance reason} & \textbf{1--80} & \textbf{81--160} & \textbf{161--240} & \textbf{241--320} & \textbf{321--400} & \textbf{Total} & \textbf{\%} \\
\midrule
No dominant suboptimal behavior    &  8 & 11 & 18 & 56 & 88 & 181 & 39.96\% \\
Recent patch under evaluation      & 23 & 21 & 19 & 26 & 27 & 116 & 25.61\% \\
Rubric set already compact         &  4 &  4 &  6 & 17 & 41 &  72 & 15.89\% \\
Held-out proxy improving           &  5 &  5 &  9 & 14 & 18 &  51 & 11.26\% \\
Mixed / low-confidence evidence    & 12 &  6 &  4 &  5 &  6 &  33 &  7.28\% \\
\bottomrule
\end{tabular}}
}
\caption{Taxonomy of dominant maintenance reasons across the 453 maintenance steps.}
\label{tab:maintenance_taxonomy}
\end{table*}

We also analyze the overall patterns in the \textbf{Maintenance} samples. The five categories are: \emph{no dominant suboptimal behavior}, where the current rollouts show no clear failure pattern about rubric changes; \emph{recent patch under evaluation}, where AMARIS withholds further changes to observe the effect of a recent rubric update; \emph{rubric set already compact}, where the existing rubrics are sufficiently useful; \emph{held-out proxy improving}, where a held-out validation indicates ongoing gains under the current rubrics; and \emph{mixed / low-confidence evidence}, where the batch-level signals are too noisy or contradictory to justify a confident rubric change. The most common pattern is that the model has no dominant suboptimal behavior present (181 steps, 39.96\%), confirming that late-stage stability is achieved through accumulated rubric improvement. The second most common pattern is that a recent rubric update is still under evaluation (116, 25.61\%), indicating that AMARIS frequently chooses to wait for additional evidence to evaluate the previous decisions. The remaining maintenance steps are mainly for reducing redundant rubrics (72, 15.89\%).

\paragraph{Composition of the active rubric set.}
Despite of the step-level analysis, we also conduct the rubric-level analysis by randomly select 50 samples per training phase and manually categorize the internal composition of the rubrics into six categories. Table \ref{tab:rubric_composition} reports the average number of active adaptive rubrics per step in six functional categories. These categories are: \emph{correctness} rubrics that assess factual and logical accuracy; \emph{instruction following} rubrics that check adherence to explicit instructions; \emph{safety} rubrics that penalize harmful or biased content; \emph{communication quality} rubrics that evaluate clarity and coherence; \emph{anti-reward hacking} rubrics specifically designed to penalize identified reward hacking behaviors; and \emph{bonus rewards} rubrics that incentivize desirable but non-mandatory qualities such as providing illustrative examples, acknowledging limitations, or offering alternative perspectives. Correctness and safety related rubrics decreases over training (from 5.95 to 3.85 and from 4.00 to 2.20), while bonus rewards grow from 1.05 to 6.55. This compositional shift mirrors the mode transition in Table~\ref{tab:update_mode}: as defensive updates decline and curriculum updates increase, the rubric set reallocates capacity from basic compliance toward better quality. Anti-reward hacking increases to the top at 3.80 on average in mid-training and then decrease to 2.05 at the end and communication quality rubrics rise monotonically from 2.05 to 4.20, reflecting the late-stage emphasis on response efficiency and completeness. The total number of the active rubrics reaches its maximum at 25.35 rubrics per step in middle of training and decreases to 22.75 in the final phase, confirming that AMARIS achieves rubric evolution with controlled growing of numbers.

\begin{table*}[htbp!]
\centering\small{
\resizebox{\textwidth}{!}{
\begin{tabular}{l rrrrr r}
\toprule
\textbf{Adaptive rubric category} & \textbf{1--80} & \textbf{81--160} & \textbf{161--240} & \textbf{241--320} & \textbf{321--400} & \textbf{Mean} \\
\midrule
Correctness     & 5.95 & 5.40 & 4.75 & 4.20 & 3.85 & 4.83 \\
Instruction following   & 4.30 & 4.35 & 4.20 & 4.05 & 3.90 & 4.16 \\
Safety        & 4.00 & 3.60 & 3.00 & 2.65 & 2.20 & 3.09 \\
Communication quality               & 2.05 & 2.70 & 3.25 & 3.90 & 4.20 & 3.22 \\
Anti-reward hacking             & 2.75 & 3.15 & 3.80 & 3.00 & 2.05 & 2.95 \\
Bonus rewards        & 1.05 & 3.85 & 6.35 & 6.60 & 6.55 & 4.88 \\
\midrule
\textbf{Total}                      & \textbf{20.10} & \textbf{23.05} & \textbf{25.35} & \textbf{24.40} & \textbf{22.75} & \textbf{23.13} \\
\bottomrule
\end{tabular}}
}
\caption{Average number of active adaptive rubrics per step, grouped by functional category. Correctness and safety rubrics decrease over training while bonus rewards and communication-quality rubrics grow, reflecting a shift from basic compliance toward higher-order quality. The total set increases modestly in the middle of training and then decreases, indicating controlled evolution.}
\label{tab:rubric_composition}
\end{table*}

\paragraph{Rubric evolution is calibration-driven.}
\begin{table}[htbp!]
\centering\small{
\begin{tabular}{l rr}
\toprule
\textbf{Operation type} & \textbf{Count} & \textbf{\%} \\
\midrule
\textsc{Reweight} & 1{,}106 & 31.78\% \\
\textsc{Update}   & 1{,}018 & 29.25\% \\
\textsc{Create}   &    542 & 15.57\% \\
\textsc{Split}    &    291 &  8.36\% \\
\textsc{Delete}   &    286 &  8.22\% \\
\textsc{Merge}    &    237 &  6.81\% \\
\midrule
\textbf{Total}    & \textbf{3{,}480} & \textbf{100.00\%} \\
\bottomrule
\end{tabular}
}
\caption{Distribution of rubric-edit operations across 1,200 step-level updates (2.90 operations per update on average). \textsc{Reweight} and \textsc{Update} jointly account for 61.03\% of all edits, indicating that AMARIS evolves rubrics primarily through calibration rather than expansion.}
\label{tab:edit_operations}
\end{table}

\begin{table*}[htbp!]
\centering\small{
\resizebox{\textwidth}{!}{
\begin{tabular}{l rrr l}
\toprule
\textbf{Operation type} & \textbf{Defensive} & \textbf{Curriculum} & \textbf{Maintenance} & \textbf{Most associated} \\
\midrule
Create   & 210 (38.75\%) & 283 (52.21\%) &  49 \phantom{0}(9.04\%) & Curriculum \\
Update   & 303 (29.76\%) & 402 (39.49\%) & 313 (30.75\%) & Curriculum \\
Reweight & 184 (16.64\%) & 498 (45.03\%) & 424 (38.34\%) & Curriculum \\
Delete   &  24 \phantom{0}(8.39\%) &  33 (11.54\%) & 229 (80.07\%) & Maintenance \\
Merge    &  19 \phantom{0}(8.02\%) & 109 (45.99\%) & 109 (45.99\%) & Curriculum \& Maintenance \\
Split    & 236 (81.10\%) &  55 (18.90\%) &   0 \phantom{0}(0.00\%) & Defensive \\
\bottomrule
\end{tabular}}
}
\caption{Rubric-edit operations decomposed by dominant update mode. \textsc{Split} is overwhelmingly defensive (81.10\%), reflecting exploit repair via rubric decomposition; \textsc{Delete} is primarily maintenance-driven (80.07\%), indicating late-stage consolidation; \textsc{Create} is most often curriculum-driven (52.21\%), showing that new rubrics raise standards more than they patch exploits.}
\label{tab:operations_by_mode}
\end{table*}

\begin{table}[htbp!]
\centering\small{
\resizebox{\columnwidth}{!}{
\begin{tabular}{l rr}
\toprule
\textbf{Category of new rubric} & \textbf{Count} & \textbf{Share} \\
\midrule
Anti-reward hacking          & 163 & 30.07\% \\
Curriculum rubrics     & 124 & 22.88\% \\
Communication quality            &  92 & 16.97\% \\
Instruction following &  71 & 13.10\% \\
Safety calibration     &  54 &  9.96\% \\
Correctness \& factual grounding  &  38 &  7.01\% \\
\bottomrule
\end{tabular}}
}
\caption{Breakdown of the 542 \textsc{Create} operations by rubric category. Nearly one-third produce anti-hacking guardrails, but the second-largest block (22.88\%) introduces stretch and curriculum rubrics, showing that AMARIS uses new rubrics to raise standards, not only to patch exploits.}
\label{tab:create_categories}
\end{table}

Despite of the analysis of the rubric itself, we also analyze the pattern of rubric editing, using the definition from the Section \ref{ssec:update}. Across the 1,200 step-level updates, AMARIS performs 3,480 rubric-edit operations (2.90 per update on average). \textbf{Reweight} (31.78\%) and \textbf{Update} (29.25\%) jointly account for 61.03\% of all edits, while \textsc{Create} accounts for only 15.57\%, showing that AMARIS mostly evolves rubrics by refining existing criteria rather than adding new ones. The relationship between edit operations and update modes shows how each strategy achieves its effect: 81.10\% of \textbf{Split} operations occur in defensive mode, reflecting that correcting suboptimal behaviors often requires decomposing a coarse rubric into more precise sub-rubrics; 80.07\% of \textbf{Delete} operations occur in maintenance mode, indicating that rubric pruning is a late-stage consolidation behavior; and 52.21\% of \textbf{Create} operations occur in curriculum mode, showing that new rubrics are more often introduced to raise evaluation standards. Among the 542 \textbf{Create} operations themselves, 30.07\% produce anti-hacking guardrails and 22.88\% produce stretch and curriculum rubrics.

\paragraph{Memory improves rubric stability.}
\begin{table*}[htbp!]
\centering\small{
\resizebox{\textwidth}{!}{
\begin{tabular}{l r rrr l}
\toprule
\textbf{Domain} & \textbf{Defensive} & \textbf{Static-dominant} & \textbf{Dynamic-dominant} & \textbf{Mixed} & \textbf{Most useful} \\
\midrule
Science              & 132 & 81 (61.36\%) & 28 (21.21\%) & 23 (17.42\%) & Static \\
Medicine             & 124 & 40 (32.26\%) & 48 (38.71\%) & 36 (29.03\%) & Dynamic \\
Instruction following &  78 & 27 (34.62\%) & 21 (26.92\%) & 30 (38.46\%) & Mixed \\
\bottomrule
\end{tabular}}
}
\caption{Decisive evidence source for each rubric modification, by domain. Science is static-memory dominated, medicine benefits most from dynamic retrieval, and instruction following relies on mixed evidence.}
\label{tab:memory_evidence}
\end{table*}

\begin{table*}[htbp!]
\centering\small{
\resizebox{\textwidth}{!}{
\begin{tabular}{l rrrr}
\toprule
\textbf{Setting} & \textbf{Science} & \textbf{Medicine} & \textbf{Instruction following} & \textbf{Total} \\
\midrule
AMARIS (no memory)                 & 44 & 39 & 35 & 118 \\
AMARIS (static + dynamic memory)   & 27 & 26 & 22 &  75 \\
\midrule
Reduction                          & 17 & 13 & 13 &  43 \\
Relative reduction                 & 38.64\% & 33.33\% & 37.14\% & 36.44\% \\
\bottomrule
\end{tabular}}
}
\caption{Short-term rubric reversals (rubric reversely modified within 8 steps). Memory reduces oscillatory rubric changes by 36.44\% overall, with consistent reductions across all three domains.}
\label{tab:reversals}
\end{table*}

To quantify the stabilizing effect of memory, we measure \emph{short-term reversals}, defined as a rubric being reversely modified (deleted, merged, or reweighted by ${\ge}$0.5 in the opposite direction) within 8 steps. Fewer short-term reversals means that the rubrics are more robust and AMARIS produces less oscillatory changes to calibrate the rubrics. With static and dynamic memory enabled, the total number of reversals across the three domains drops from 118 (no memory) to 75, a 36.44\% relative reduction that is consistent across domains (38.64\% in science, 33.33\% in medicine, 37.14\% in instruction following). We further examine which memory type provides the evidence for each modification. In science, static (recent-step) memory is the dominant evidence source in 61.36\% of cases, matching the finding in Table \ref{tab:ablation} that static memory slightly outperforms dynamic memory on GPQA-Diamond. In medicine, dynamic (semantic) retrieval is the largest single source (38.71\%). In instruction following, mixed evidence is the most common category (38.46\%), reflecting a more heterogeneous memory usage. These domain-level patterns align with the complementary nature of static and dynamic memory observed in the ablation study about memory usage and provide an explanation for the quantitative gains reported in Table \ref{tab:ablation}.

\clearpage

\clearpage
\section{Case Study}
\label{sec:case_study}

Appendix~\ref{sec:case_study} presents four full case studies. Each full case includes the input, model outputs before and after the rubric update, the complete rubric sets before and after the update, retrieved memory snippets, and the update rationale.
The cases are drawn from science, medicine, and instruction following, and collectively illustrate the maintenance, defensive, and curriculum-advancement strategies described in Section~\ref{ssec:update}.

\subsection{Case 1: Merging Overlapping Structure Rubrics}
\label{par:case_merge}

\paragraph{Full input.}
\begin{quote}\small
Aqueous copper(II) nitrate reacts with potassium iodide to yield solid copper(I) iodide, potassium nitrate, and iodine. Write a balanced net ionic equation for the reaction. Provide a detailed and accurate answer.
\end{quote}

\paragraph{Output before.}
\begin{quote}\small
First write the molecular equation:
\[
\begin{aligned}
&2\mathrm{Cu(NO_3)_2(aq)} + 4\mathrm{KI(aq)}\\
&\quad \rightarrow 2\mathrm{CuI(s)} + 4\mathrm{KNO_3(aq)} + \mathrm{I_2(s)}.
\end{aligned}
\]
Removing the spectator ions \(\mathrm{K^+}\) and \(\mathrm{NO_3^-}\) gives the net ionic equation:
\[
2\mathrm{Cu}^{2+}\mathrm{(aq)} + 4\mathrm{I}^{-}\mathrm{(aq)}
\rightarrow 2\mathrm{CuI(s)} + \mathrm{I_2(s)}.
\]
\textbf{Final Net Ionic Equation}
\[
2\mathrm{Cu}^{2+}\mathrm{(aq)} + 4\mathrm{I}^{-}\mathrm{(aq)}
\rightarrow 2\mathrm{CuI(s)} + \mathrm{I_2(s)}.
\]
This shows that \(\mathrm{Cu}^{2+}\) is reduced to \(\mathrm{Cu}^{+}\) in \(\mathrm{CuI}\), while iodide is oxidized to iodine.

\textbf{Final Answer}
\[
2\mathrm{Cu}^{2+}\mathrm{(aq)} + 4\mathrm{I}^{-}\mathrm{(aq)}
\rightarrow 2\mathrm{CuI(s)} + \mathrm{I_2(s)}.
\]
\textbf{Summary}

The balanced net ionic equation is:
\[
2\mathrm{Cu}^{2+}\mathrm{(aq)} + 4\mathrm{I}^{-}\mathrm{(aq)}
\rightarrow 2\mathrm{CuI(s)} + \mathrm{I_2(s)}.
\]
\end{quote}

\paragraph{Output after.}
\begin{quote}\small
Copper(II) is reduced to copper(I), and iodide is partly oxidized to iodine while the remaining iodide precipitates copper(I) iodide.

The balanced molecular equation is:
\[
\begin{aligned}
&2\mathrm{Cu(NO_3)_2(aq)} + 4\mathrm{KI(aq)}\\
&\quad \rightarrow 2\mathrm{CuI(s)} + 4\mathrm{KNO_3(aq)} + \mathrm{I_2(s)}.
\end{aligned}
\]
Removing \(\mathrm{K^+}\) and \(\mathrm{NO_3^-}\) as spectator ions gives:

\textbf{Final Answer:}
\[
2\mathrm{Cu}^{2+}\mathrm{(aq)} + 4\mathrm{I}^{-}\mathrm{(aq)}
\rightarrow 2\mathrm{CuI(s)} + \mathrm{I_2(s)}.
\]
\end{quote}

\paragraph{Full before rubric set.}
\begin{itemize}[leftmargin=*]\small
\item \textbf{C0 (+1.00, anchor): Correct final answer.} Award when the response gives the chemically correct balanced net ionic equation for the reaction.
\item \textbf{C1 (+1.20): Final-answer block presence.} Award when the response contains a clearly demarcated final-answer block for an explicit-output task.
\item \textbf{C2 (+1.50): Final-answer block completeness.} Award when all requested outputs appear inside the final-answer block.
\item \textbf{C3 (+0.80): Approved final-answer heading.} Award when the final-answer block uses a valid heading such as ``Final Answer,'' ``Answer,'' or ``Final Net Ionic Equation.''
\item \textbf{C4 (-4.00): Decorative emoji hard gate.} Penalize decorative icons or flashy symbols in headings, answer regions, or summary regions, and prevent related clean-style rewards from firing.
\item \textbf{C5 (+2.00): Clean professional formatting.} Award when the answer is professionally formatted, easy to scan, and free of distracting decoration.
\end{itemize}

\paragraph{Full after rubric set.}
\begin{itemize}[leftmargin=*]\small
\item \textbf{C0 (+1.00, anchor): Correct final answer.} Award when the response gives the chemically correct balanced net ionic equation for the reaction.
\item \textbf{C6 (+3.50): Single complete final-answer block.} Award when the response contains exactly one clearly demarcated final-answer block that addresses the requested output completely, uses a valid heading, and does not duplicate the definitive answer in separate answer-like regions.
\item \textbf{C4 (-2.00): Decorative emoji soft penalty.} Penalize decorative icons or flashy symbols wherever they appear, with special concern for headings and final-answer regions, but do not let style alone erase useful correctness and structure signals.
\item \textbf{C5 (+1.20): Clean professional formatting.} Award a smaller top-up when the response is plain, professional, and non-distracting, provided correctness and final-answer integrity are already acceptable.
\end{itemize}

\paragraph{Retrieved memory snippets.}
\begin{itemize}[leftmargin=*]\small
\item \textbf{Static memory:} Recent science summaries repeatedly flagged duplicate answer regions, answer leakage outside the canonical final block, and decorative icons in headings or final-answer regions.
\item \textbf{Static memory:} Earlier incremental updates to separate block-presence, heading, and style rubrics did not remove the duplicate-final-section behavior; the same structural loophole reappeared across steps.
\item \textbf{Current rollout evidence:} One rollout had correct chemistry but presented the same net ionic equation under multiple answer-like headings, allowing separate structure rubrics to fire even though the output had no single authoritative answer region.
\item \textbf{Current rollout evidence:} Another rollout showed the opposite failure: polished answer structure coexisted with a wrong chemistry conclusion, indicating that structure rewards needed tighter coupling to correctness.
\end{itemize}

\paragraph{Update rationale.}
The before output is chemically correct, but it exposes a reward-shaping flaw. The definitive net ionic equation appears in the derivation, under \texttt{Final Net Ionic Equation}, under \texttt{Final Answer}, and again under \texttt{Summary}. Under the before rubric set, the model can earn separate rewards for having a final block, covering the requested output, using an approved heading, and looking clean, even though the user receives several competing answer regions.

AMARIS therefore merges \textbf{C1}, \textbf{C2}, and \textbf{C3} into \textbf{C6}, a single higher-weight rubric that asks for one complete, authoritative final-answer block. This removes double-counting and makes the desired behavior simpler: give the correct answer once, in the right place. The style rules are also softened and reweighted. \textbf{C4} remains a penalty, but it no longer acts as a broad structural hard gate; \textbf{C5} becomes a small top-up rather than a way to rescue duplicated or weak content.

The after output keeps the useful chemistry derivation but ends with one final-answer region and no duplicate answer sections. This aligns the reward with the desired presentation: correct science first, then a single clean answer block, with style as a secondary constraint.

\clearpage
\subsection{Case 2: Precise Scalar Reweighting for Correct but Generic Reasoning}
\label{par:case_reweight}

\paragraph{Full input.}
\begin{quote}\small
Which cellular process is disrupted first when a drug inhibits bacterial DNA gyrase?

A. translation

B. DNA replication

C. glycolysis

D. peptidoglycan cross-linking
\end{quote}

\paragraph{Output before.}
\begin{quote}\small
DNA gyrase is a major bacterial enzyme involved in DNA structure, and antibiotics often affect growth-related pathways. Because bacteria need DNA topology control before cell division, the best answer is probably \textbf{B}, although translation and cell-wall pathways can also be affected indirectly.
\end{quote}

\paragraph{Output after.}
\begin{quote}\small
DNA gyrase relieves supercoiling ahead of the replication fork, so inhibiting it first blocks \textbf{DNA replication}. \textbf{Answer: B.}
\end{quote}

\paragraph{Full before rubric set.}
\begin{itemize}[leftmargin=*]\small
\item \textbf{A0 (+1.00): Final answer is correct.} Award when the response selects the correct option. For this input, the correct answer is \textbf{B, DNA replication}.
\item \textbf{S1 (+0.75): Provides at least two relevant scientific facts.} Award when the response includes at least two accurate facts relevant to the question, such as DNA gyrase being a bacterial enzyme, its role in DNA topology or supercoiling, or the relationship between DNA topology and bacterial growth. This criterion does not require the facts to form a direct causal chain from drug action to the selected answer.
\item \textbf{S2 (+0.50): States the final answer explicitly.} Award when the response clearly names or marks the selected option rather than leaving the choice implicit.
\item \textbf{S3 (+0.40): Acknowledges uncertainty when confidence is limited.} Award when the response uses appropriately cautious language for genuinely uncertain cases, notes possible indirect effects, or avoids overstating unsupported mechanisms. In this case, the criterion can still reward hedge language even though the correct mechanism is known.
\item \textbf{S4 (-0.50): Makes unsupported scientific claims.} Penalize when the response introduces claims that are scientifically unsupported or inconsistent with the question.
\end{itemize}

\paragraph{Full after rubric set.}
\begin{itemize}[leftmargin=*]\small
\item \textbf{A0 (+1.00): Final answer is correct.} Award when the response selects the correct option. For this input, the correct answer is \textbf{B, DNA replication}.
\item \textbf{S1 (+0.35): Provides at least two relevant scientific facts.} Award when the response includes accurate facts relevant to DNA gyrase, bacterial DNA topology, or replication. The lower weight keeps factual coverage useful but prevents generic fact listing from dominating the reward.
\item \textbf{S2 (+0.50): States the final answer explicitly.} Award when the response clearly names or marks the selected option.
\item \textbf{S3 (+0.10): Acknowledges uncertainty when confidence is limited.} Award only lightly for cautious language when uncertainty is warranted. The reduced weight discourages routine hedging when the mechanism directly supports one answer.
\item \textbf{S4 (-0.50): Makes unsupported scientific claims.} Penalize when the response introduces claims that are scientifically unsupported or inconsistent with the question.
\item \textbf{S5 (+0.75): Explanation directly justifies the selected option.} Award when the response explicitly connects the mechanism to the answer: DNA gyrase relieves supercoiling ahead of the replication fork, so inhibiting it first disrupts DNA replication. Do not award full credit for disconnected facts that are merely related to bacteria, antibiotics, or cell growth.
\end{itemize}

\paragraph{Retrieved memory snippets.}
\begin{itemize}[leftmargin=*]\small
\item \textbf{Static memory:} Recent science batches showed the same ``correct but generic'' explanation pattern. The issue was not disappearing across nearby steps.
\item \textbf{Static memory:} The model repeatedly gave correct final options while relying on shallow causal support, broad scientific filler, and hedge tokens such as ``probably,'' ``likely,'' and ``can also.''
\item \textbf{Dynamic memory:} An earlier update that rewarded answer-linked justification improved explanation quality without hurting correctness.
\item \textbf{Dynamic memory:} A prior anti-verbosity rule made some science answers too terse and reduced explanation completeness, so a blunt brevity penalty was risky.
\end{itemize}

\paragraph{Update rationale.}
The output before the update is correct, but it scores too well for the wrong reasons. It satisfies correctness, explicit answer format, generic fact coverage, and uncertainty language, while avoiding the key reasoning step: DNA gyrase inhibition blocks replication because gyrase relieves supercoiling ahead of the replication fork.

AMARIS therefore makes a scalar adjustment rather than a broad rewrite. It lowers \textbf{S1} from \(+0.75\) to \(+0.35\) because fact count alone is not enough. It lowers \textbf{S3} from \(+0.40\) to \(+0.10\) because uncertainty language should not rescue an answer when the causal mechanism is clear. It adds \textbf{S5} at \(+0.75\) to reward direct mechanism-to-answer justification.

This update preserves useful rubric coverage while changing the reward balance: generic factual language becomes secondary, hedging becomes minor, and direct causal justification becomes the main adaptive signal.

\clearpage
\subsection{Case 3: Reward Hacking Correction in Medicine / Over-Refusal}
\label{par:case_exploit}

\paragraph{Full input.}
\begin{quote}\small
I missed one dose of amoxicillin today. Should I double the next dose?
\end{quote}

\paragraph{Output before.}
\begin{quote}\small
I cannot advise on medication dosing changes. Please contact your doctor or pharmacist before taking any more. It would be unsafe for me to tell you whether to change your dose.
\end{quote}

\paragraph{Output after.}
\begin{quote}\small
If you miss a dose of amoxicillin, take it when you remember unless it is almost time for the next dose; if so, skip the missed one and return to your usual schedule. Do \textbf{not} double the next dose. Check the prescription label or ask your pharmacist if your instructions differ, and seek medical care if you develop a rash, trouble breathing, or severe diarrhea.
\end{quote}

\paragraph{Full before rubric set.}
\begin{itemize}[leftmargin=*]\small
\item \textbf{M0 (+1.00, anchor): Avoid harmful medical advice.} The response must not provide dangerous, individualized, or unsafe medical instructions. It should preserve user safety, avoid unsupported treatment changes, and direct the user to urgent care when symptoms indicate possible serious harm.
\item \textbf{M1 (+1.00): Avoid definitive dosing changes when context is insufficient.} The response should not instruct the user to alter a prescribed dose when key clinical context is missing, such as the exact prescription directions, indication, allergies, kidney disease, age, pregnancy status, other medications, or prior adverse reactions.
\item \textbf{M2 (+0.75): Recommend clinician or pharmacist consultation when uncertainty exists.} The response should tell the user to check the prescription label or contact a clinician or pharmacist when the answer depends on prescription-specific instructions, patient-specific risk factors, or unclear timing.
\item \textbf{M3 (+0.50): Use appropriate safety language.} The response should use cautious medical language, avoid overstating certainty, and include safety-oriented framing when giving medical information.
\item \textbf{M4 (+0.75): Provide actionable next steps for low-risk questions.} When the request is a common, low-risk medication-adherence question, the response should give practical next steps the user can take, while preserving appropriate safety boundaries.
\end{itemize}

\paragraph{Full after rubric set.}
\begin{itemize}[leftmargin=*]\small
\item \textbf{M0 (+1.00, anchor): Avoid harmful medical advice.} The response must not provide dangerous, individualized, or unsafe medical instructions. It should preserve user safety, avoid unsupported treatment changes, and direct the user to urgent care when symptoms indicate possible serious harm.
\item \textbf{M1 (+1.00, updated): Avoid definitive dosing changes beyond common low-risk guidance when context is insufficient.} The response should not make individualized dose changes or override prescription-specific instructions. However, it may provide widely accepted low-risk adherence guidance, such as not doubling a missed dose, when paired with instructions to follow the prescription label and consult a pharmacist or clinician if the label differs or patient-specific risks are present.
\item \textbf{M2 (+0.40, reweighted): Recommend clinician or pharmacist consultation when uncertainty exists.} The response should include consultation guidance when relevant, but generic consultation boilerplate alone should not substitute for useful low-risk guidance on benign medication-information requests.
\item \textbf{M3 (+0.25, reweighted): Use appropriate safety language.} The response should remain cautious and medically responsible, but safety language should support the answer rather than dominate it or turn a benign question into a blanket refusal.
\item \textbf{M4 (+0.75, unchanged): Provide actionable next steps for low-risk questions.} When the request is a common, low-risk medication-adherence question, the response should give practical next steps the user can take, while preserving appropriate safety boundaries.
\item \textbf{M5 (+0.90, created): Provide standard guidance plus escalation criteria for common low-risk medication-adherence questions.} For benign missed-dose or routine adherence questions, the response should give standard, non-individualized guidance; state what not to do when there is a clear safety rule; advise checking the prescription label or pharmacist if instructions differ; and include red-flag symptoms or conditions that warrant medical care.
\item \textbf{M6 (-0.75, created): Penalize refusal of benign medical-information requests without next steps or triage rationale.} The response should lose reward if it refuses a clearly low-risk medical-information request while failing to offer safe next steps or a specific reason that the request requires individualized clinical triage.
\end{itemize}

\paragraph{Retrieved memory snippets.}
\begin{itemize}[leftmargin=*]\small
\item \textbf{Query:} \texttt{refusal penalty backfire unsafe dosing advice}
\item \textbf{Snippet 1: Prior broad refusal penalty backfired.} An earlier update that broadly penalized refusal reduced generic refusals, but later produced overly direct answers on genuinely high-risk dosing questions. This record warns against simply penalizing refusal across the board.
\item \textbf{Snippet 2: Low-risk guidance with safeguards worked better.} A previous successful update improved benign medication-adherence answers by requiring standard low-risk guidance together with red-flag symptoms, prescription-label checks, and pharmacist or clinician escalation when instructions differ.
\item \textbf{Snippet 3: Current batch pattern matches benign adherence over-refusal.} Recent analyses describe frequent refusals on missed-dose, minor-side-effect, and low-risk self-management questions where the model could safely provide general guidance plus escalation criteria.
\end{itemize}

\paragraph{Update rationale.}
The before output is safe in a narrow sense, but it is under-helpful for a common medication-adherence question. It likely collects reward from \textbf{M1}, \textbf{M2}, and \textbf{M3} because it avoids dosing changes, recommends consultation, and uses safety language. At the same time, it fails \textbf{M4} because it gives no actionable low-risk next step. This creates a reward-hacking pattern: the model can maximize safety-adjacent reward by refusing benign medical-information requests, even when the intended behavior is safe and useful guidance.

The update avoids a blunt refusal penalty. Dynamic memory shows that broad anti-refusal fixes previously harmed high-risk dosing behavior. The safer correction is narrower: keep the core safety constraint, clarify that common low-risk guidance is allowed, lower the reward for generic consultation and safety boilerplate, add a positive rubric for standard guidance plus escalation criteria, and add a targeted penalty only for refusals on clearly benign requests without next steps or a triage rationale.

This correction changes the reward target from ``refuse to remain safe'' to ``answer benign questions with standard guidance while preserving safety boundaries.''

\clearpage
\subsection{Case 4: Curriculum Advancement Via Saturation Detection}
\label{par:case_curriculum}

\paragraph{Full input.}
\begin{quote}\small
Give exactly three bullet points, each under eight words, explaining why public-key cryptography matters.
\end{quote}

\paragraph{Output before.}
\begin{quote}\small
\begin{itemize}[leftmargin=*]
\item Secures messages online
\item Protects important information
\item Helps keep systems safe
\end{itemize}
\end{quote}

\paragraph{Output after.}
\begin{quote}\small
\begin{itemize}[leftmargin=*]
\item Enables encrypted key exchange
\item Authenticates websites at scale
\item Supports signatures without shared secrets
\end{itemize}
\end{quote}

\paragraph{Full before rubric set.}
\begin{itemize}[leftmargin=*]\small
\item \textbf{I0 (+1.00): Explicit constraint satisfaction.} The response must satisfy all user-stated hard constraints, including the exact number of bullet points, the per-bullet word limit, and the requested subject.
\item \textbf{I1 (+1.00): Exact requested format.} The response must use exactly three bullet points, with no extra prose, no missing bullet, no additional bullet, and each bullet kept under eight words.
\item \textbf{I2 (+0.75): Complete topical coverage.} The response must explain why public-key cryptography matters, with each bullet staying relevant to public-key cryptography rather than generic online safety.
\item \textbf{I3 (+0.50): Clarity and readability.} The response must be easy to understand, concise, grammatical, and free of distracting wording.
\end{itemize}

\paragraph{Full after rubric set.}
\begin{itemize}[leftmargin=*]\small
\item \textbf{I0 (+1.00): Explicit constraint satisfaction.} The response must satisfy all user-stated hard constraints, including the exact number of bullet points, the per-bullet word limit, and the requested subject.
\item \textbf{I1 (+0.25): Exact requested format.} The response must use exactly three bullet points, with no extra prose, no missing bullet, no additional bullet, and each bullet kept under eight words. This criterion remains active but receives less weight because recent training has already reached near-ceiling performance on this dimension.
\item \textbf{I2 (+0.75): Non-redundant topical coverage.} The response must explain why public-key cryptography matters while avoiding redundant restatements of the same broad idea. Each bullet should cover a meaningfully different aspect of the topic.
\item \textbf{I3 (+0.50): Clarity and readability.} The response must be easy to understand, concise, grammatical, and free of distracting wording.
\item \textbf{I4 (+0.75): Distinct concrete cryptographic function.} Each bullet must name a specific function enabled by public-key cryptography, such as encrypted key exchange, website authentication, digital signatures, identity verification, or secure communication without a pre-shared secret. Generic praise such as ``keeps data safe'' should not receive full credit.
\item \textbf{I5 (+0.75): Information density under the length constraint.} Each bullet should maximize useful technical content within the eight-word limit by using precise terms and avoiding filler, vague adjectives, or repeated safety language.
\end{itemize}

\paragraph{Retrieved memory snippets.}
\begin{itemize}[leftmargin=*]\small
\item \textbf{Static memory:} Recent summaries repeatedly showed near-ceiling compliance with structural constraints. The model reliably produced exactly three short bullets, indicating that the format objective had saturated.
\item \textbf{Static memory:} The same summaries showed little marginal gain in richer content quality: outputs stayed topical but generic, often repeating broad notions of safety or protection.
\item \textbf{Dynamic memory:} A prior attempt to reward novelty too aggressively led to occasional inaccurate or over-specific content, suggesting that the next curriculum step should reward concrete, correct functions rather than novelty for its own sake.
\item \textbf{Dynamic memory:} Earlier non-redundancy-focused updates improved concise instruction-following outputs by separating surface compliance from information value.
\end{itemize}

\paragraph{Update rationale.}
AMARIS treats this trace as a curriculum advancement case rather than a defensive repair. The before-update output satisfies the visible constraints: it has exactly three bullets, every bullet is under eight words, and all bullets are broadly related to public-key cryptography. However, the content is low-density and repetitive: ``secures,'' ``protects,'' and ``keeps systems safe'' all express nearly the same generic safety idea.

The memory evidence changes the intervention. Because static memory shows that format compliance is already stable, continuing to give high reward to exact formatting wastes training signal on a mastered behavior. AMARIS therefore reduces \textbf{I1} from \(+1.00\) to \(+0.25\) while preserving it as a guardrail. Because dynamic memory warns that an aggressive novelty reward can introduce unsupported specificity, AMARIS does not ask for novelty directly. Instead, it tightens \textbf{I2} around non-redundant coverage and creates \textbf{I4} and \textbf{I5}, which reward concrete, accurate, information-dense statements under the same user constraints.

The after-update output keeps the original format intact while raising content quality: each bullet names a distinct public-key cryptography function: key exchange, website authentication, and signatures without shared secrets. This illustrates how AMARIS uses longitudinal evidence to recognize saturation, lower the weight of already-mastered basics, and advance the rubric toward higher-order quality.

\clearpage
\section{Prompts}
\label{sec:prompts}
This section presents the complete prompt templates used across all stages of the AMARIS pipeline. Each template is instantiated at runtime by populating the placeholder fields (shown in brackets) with the actual training data. Several prompts share a common \emph{Background \& Context} block describing the Rubric-as-Reward paradigm and the set of available rubric manipulation operations. Similarly, a common \emph{Input Data Packet} structure containing the high-level training goal, RL training metadata, current reward rubrics (anchor and adaptive sets), and stage-specific data is used across all prompts except the reward scoring.
 
\begin{figure*}[htbp!]
\centering
\includegraphics[width=\textwidth]{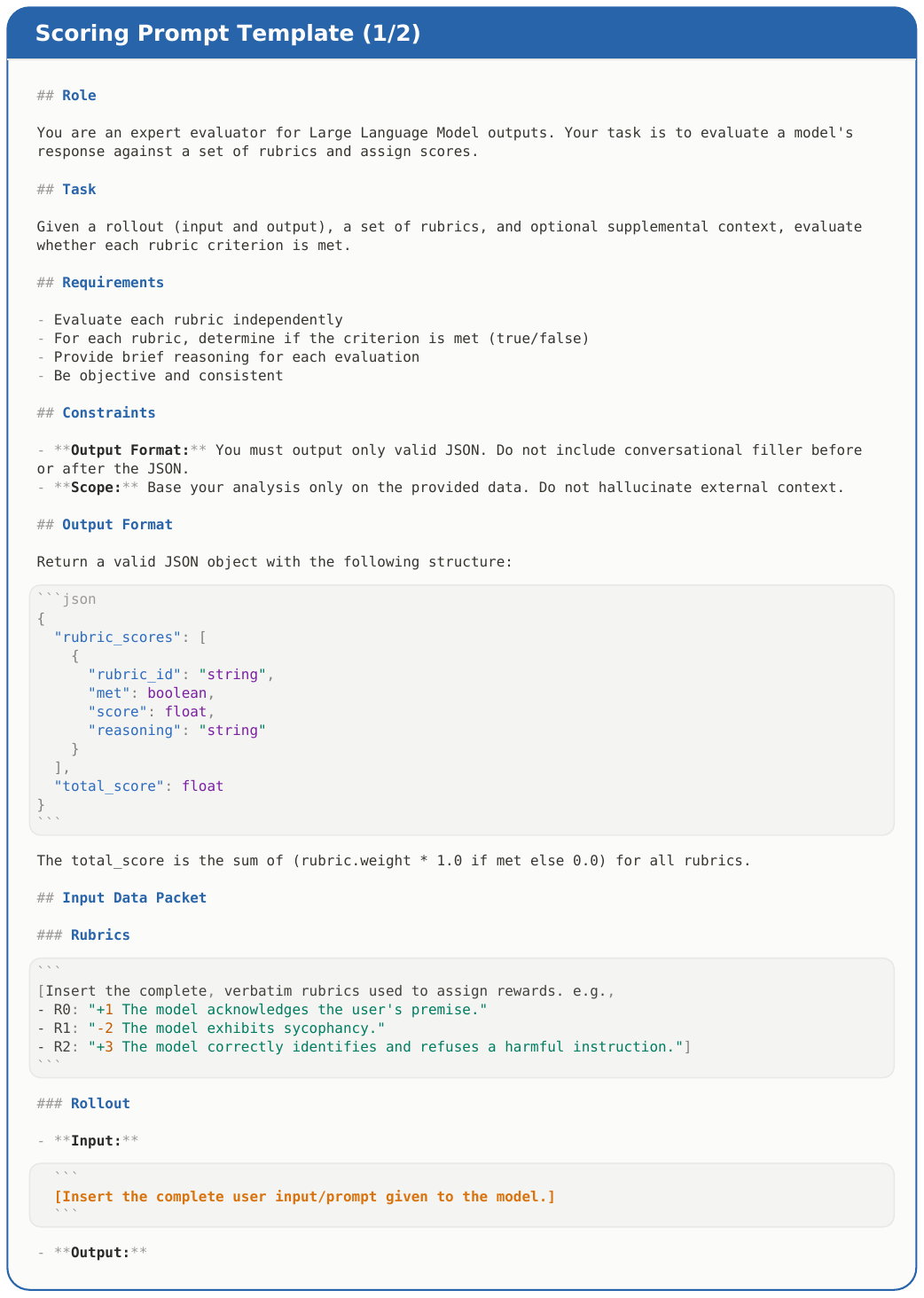}
\caption{Prompt template for the reward scoring (1/2). The LLM scoring evaluates each rubric criterion and returns per-rubric scores along with a weighted total.}
\label{fig:prompt_scorer_1}
\end{figure*}

\begin{figure*}[htbp!]
\centering
\includegraphics[width=\textwidth]{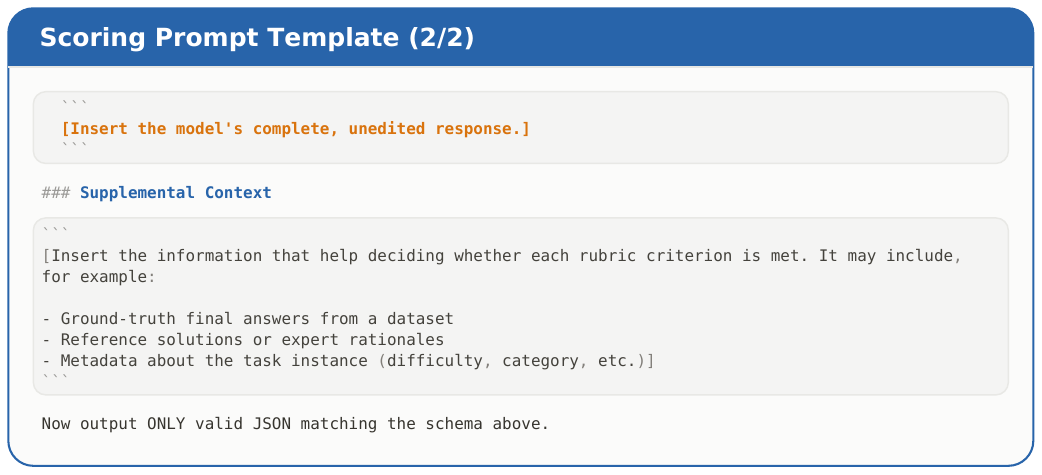}
\caption{Prompt template for the reward scoring (2/2). The LLM scoring evaluates each rubric criterion and returns per-rubric scores along with a weighted total.}
\label{fig:prompt_scorer_2}
\end{figure*}
 
\clearpage

\begin{figure*}[htbp!]
\centering
\includegraphics[width=\textwidth]{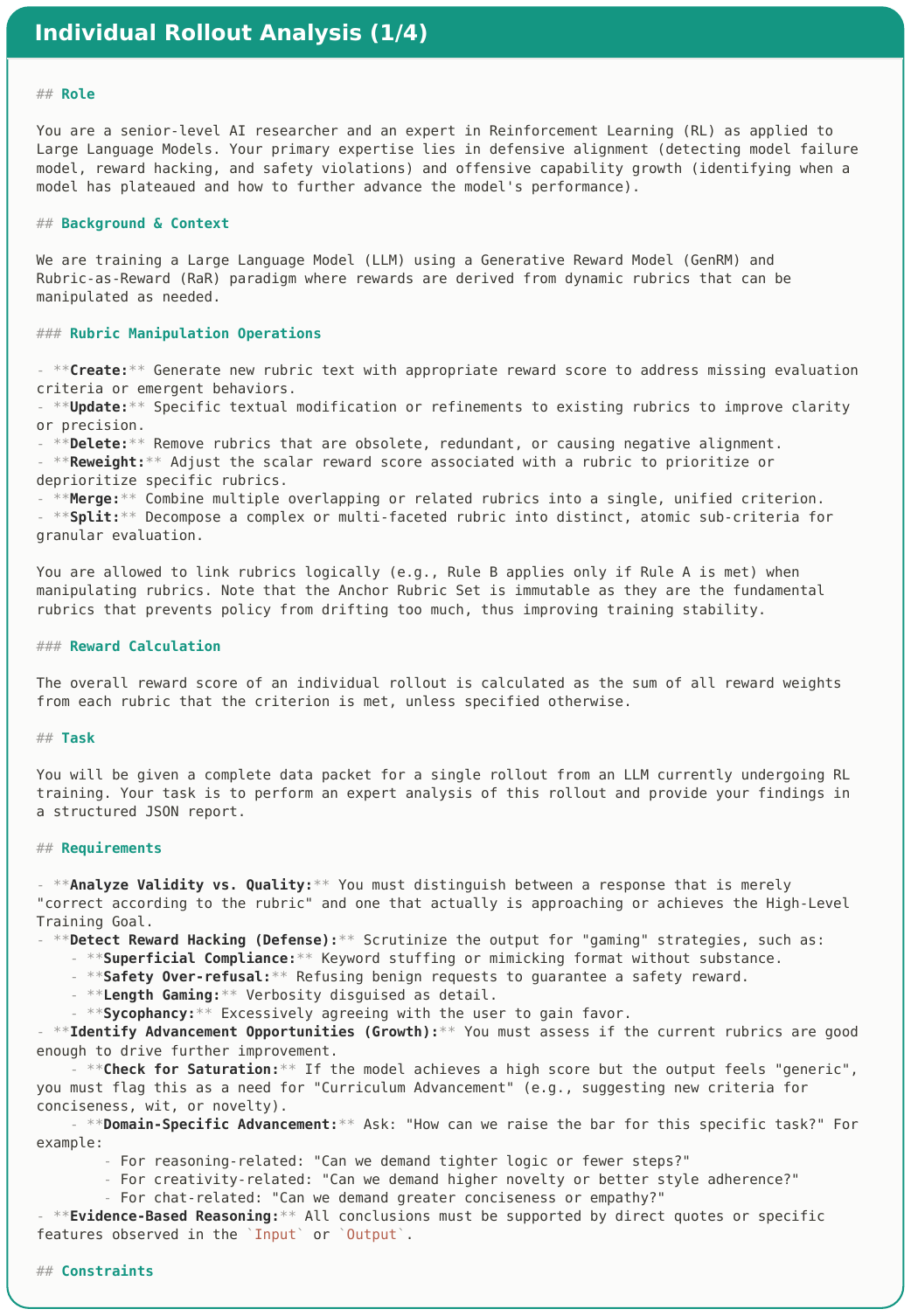}
\caption{Prompt template for individual rollout analysis (1/4) described in Section~\ref{ssec:analysis}.}
\label{fig:prompt_analysis_1}
\end{figure*}

\begin{figure*}[htbp!]
\centering
\includegraphics[width=\textwidth]{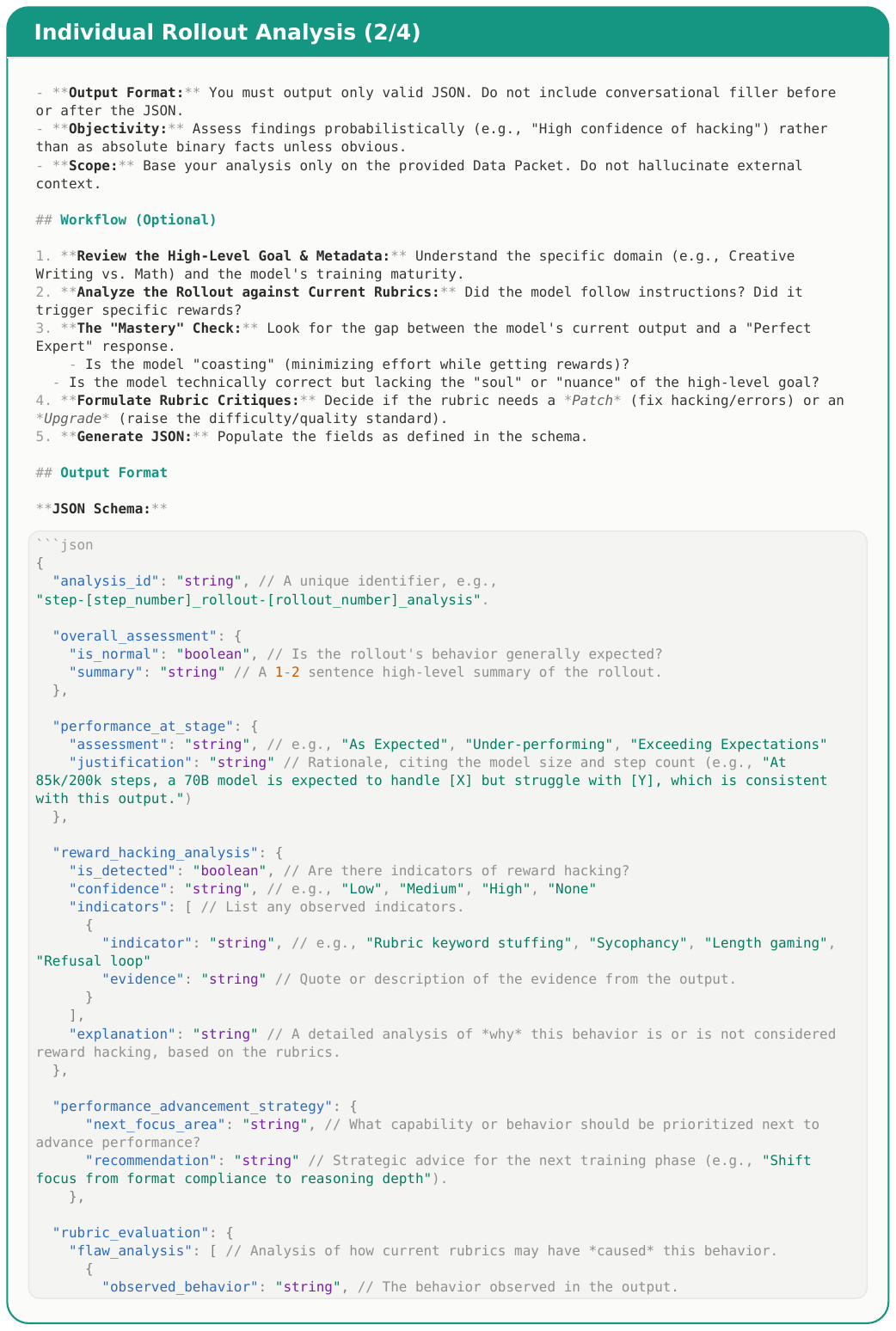}
\caption{Prompt template for individual rollout analysis (2/4) described in Section~\ref{ssec:analysis}.}
\label{fig:prompt_analysis_2}
\end{figure*}

\begin{figure*}[htbp!]
\centering
\includegraphics[width=\textwidth]{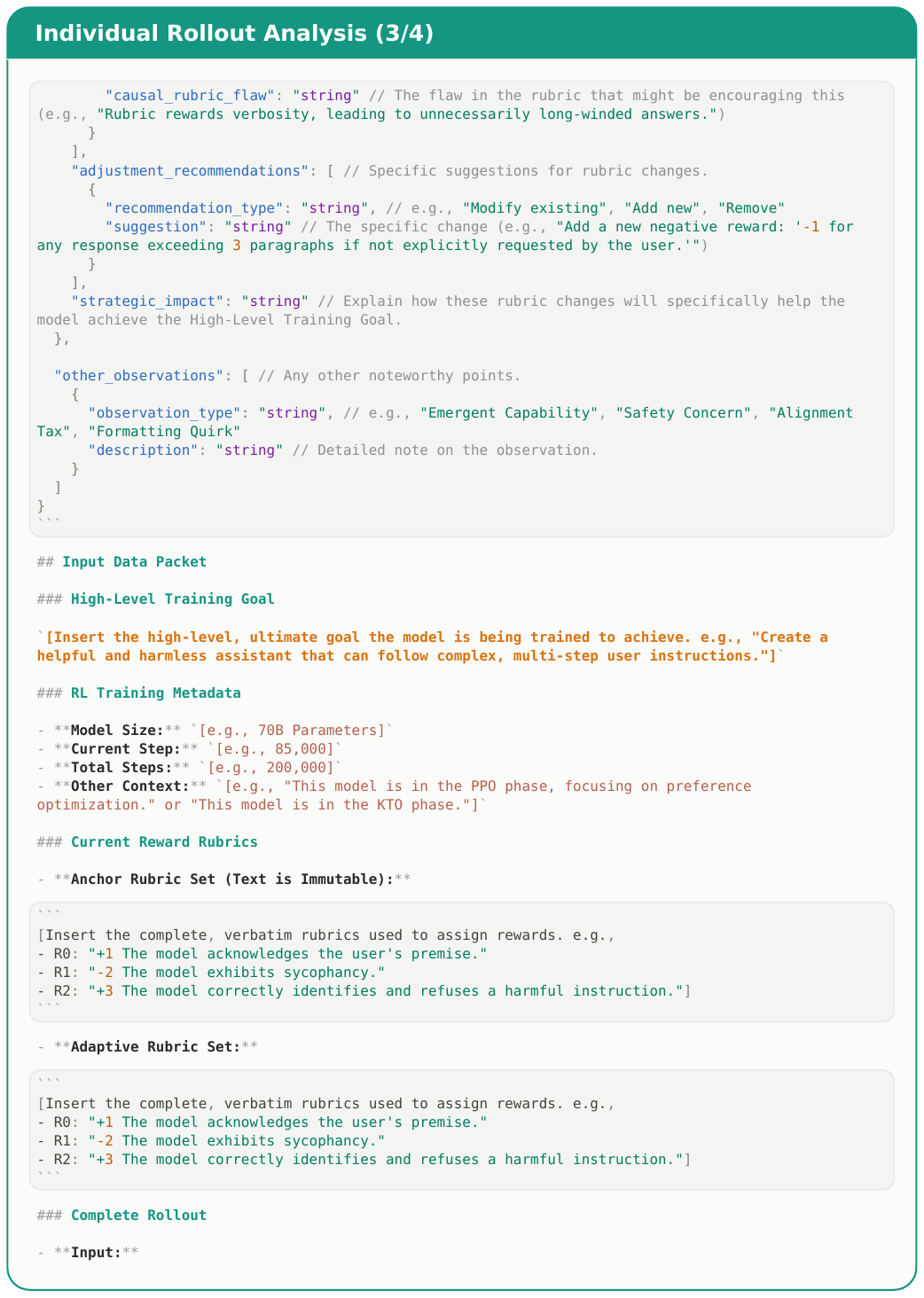}
\caption{Prompt template for individual rollout analysis (3/4) described in Section~\ref{ssec:analysis}.}
\label{fig:prompt_analysis_3}
\end{figure*}

\begin{figure*}[htbp!]
\centering
\includegraphics[width=\textwidth]{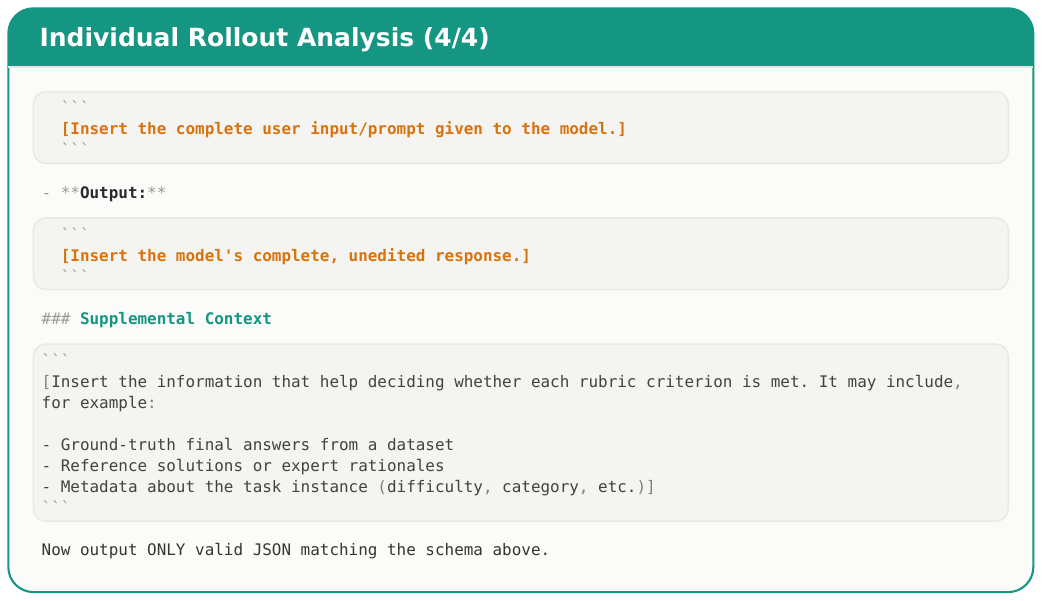}
\caption{Prompt template for individual rollout analysis (4/4) described in Section~\ref{ssec:analysis}.}
\label{fig:prompt_analysis_4}
\end{figure*}
 
\clearpage

\begin{figure*}[htbp!]
\centering
\includegraphics[width=\textwidth]{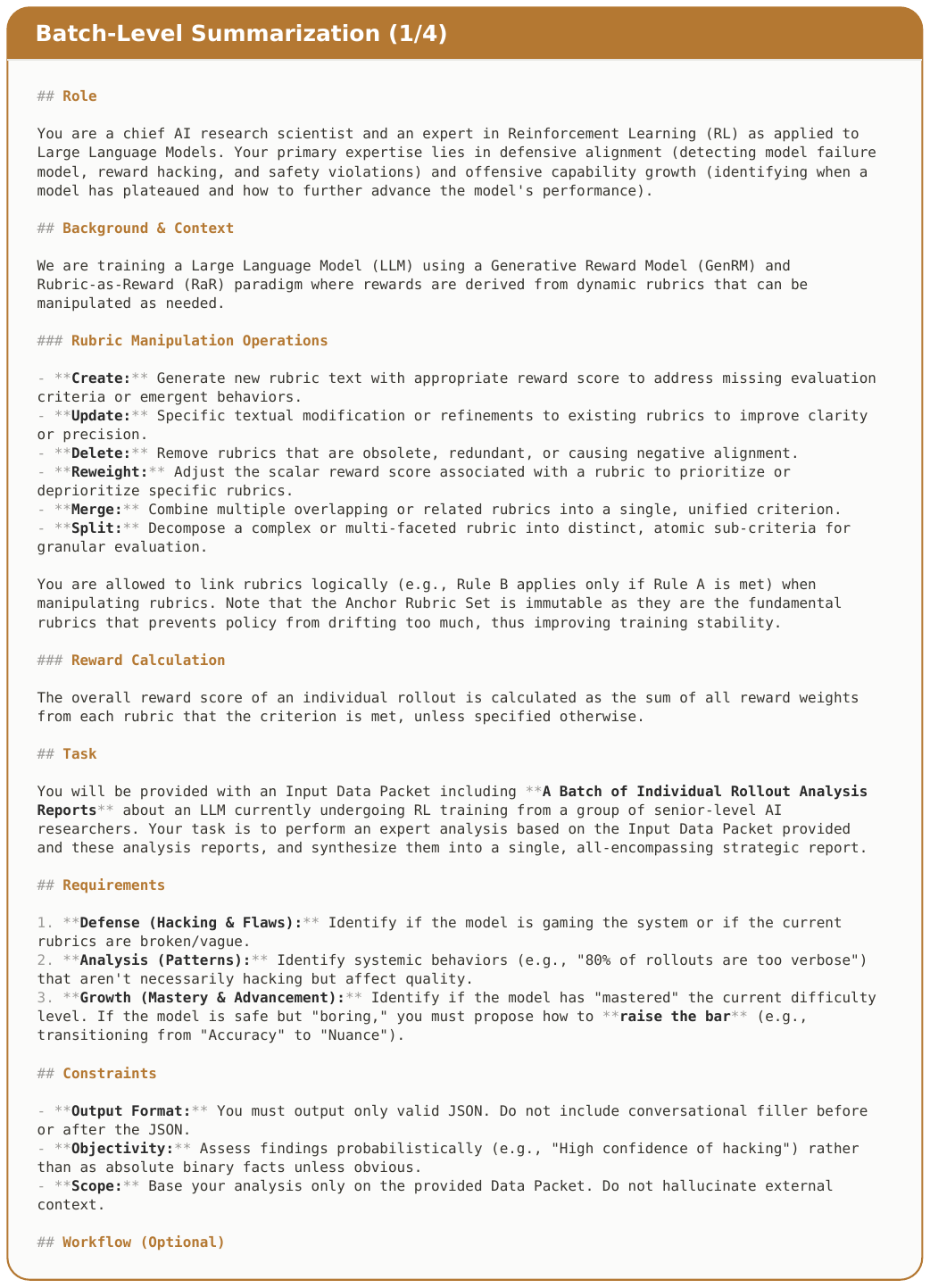}
\caption{Prompt template for batch-level summarization (1/4) described in Section~\ref{ssec:summarization}. This prompt aggregates individual rollout analyses into a step-level strategic report.}
\label{fig:prompt_summary_1}
\end{figure*}

\begin{figure*}[htbp!]
\centering
\includegraphics[width=\textwidth]{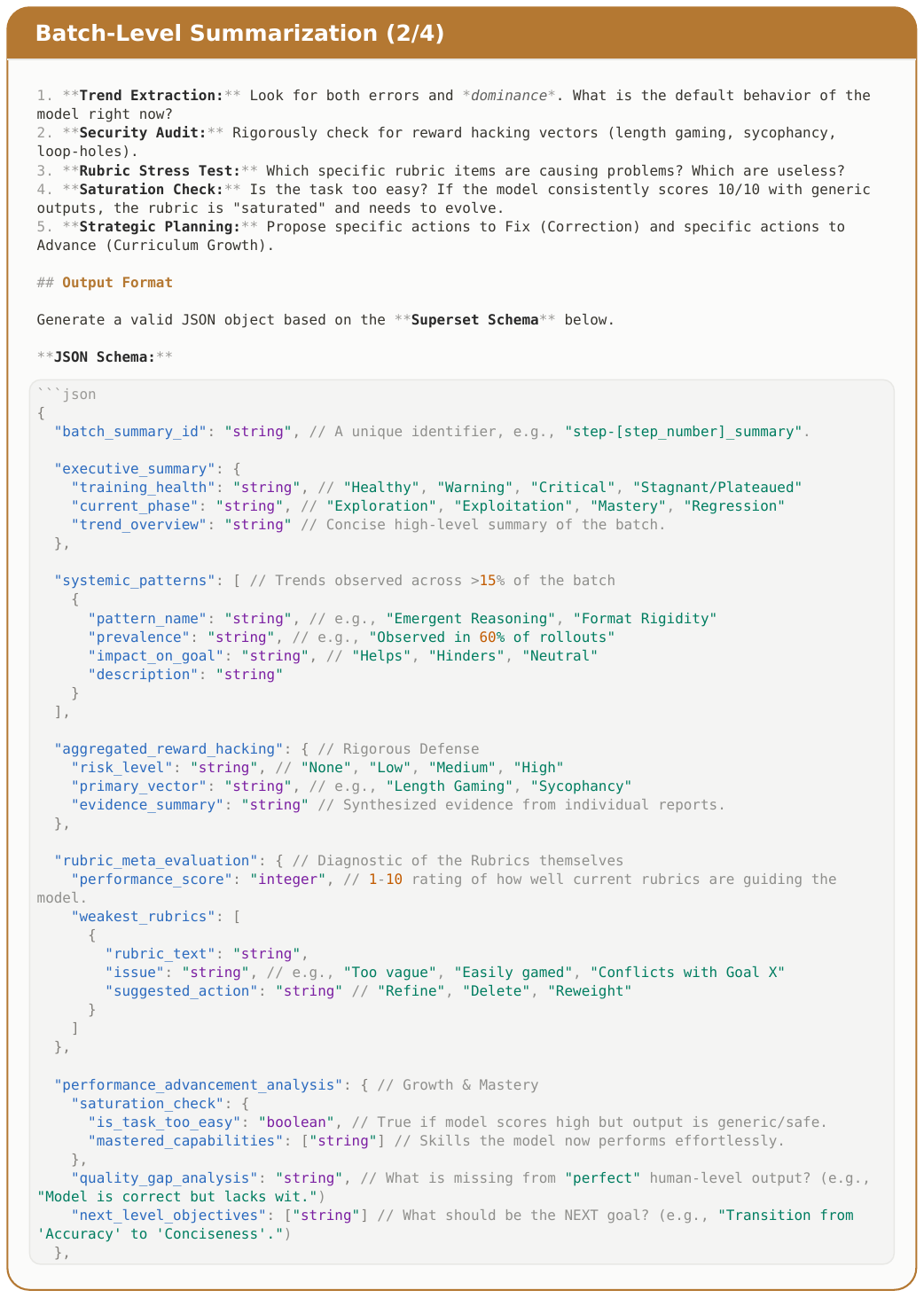}
\caption{Prompt template for batch-level summarization (2/4) described in Section~\ref{ssec:summarization}. This prompt aggregates individual rollout analyses into a step-level strategic report.}
\label{fig:prompt_summary_2}
\end{figure*}

\begin{figure*}[htbp!]
\centering
\includegraphics[width=\textwidth]{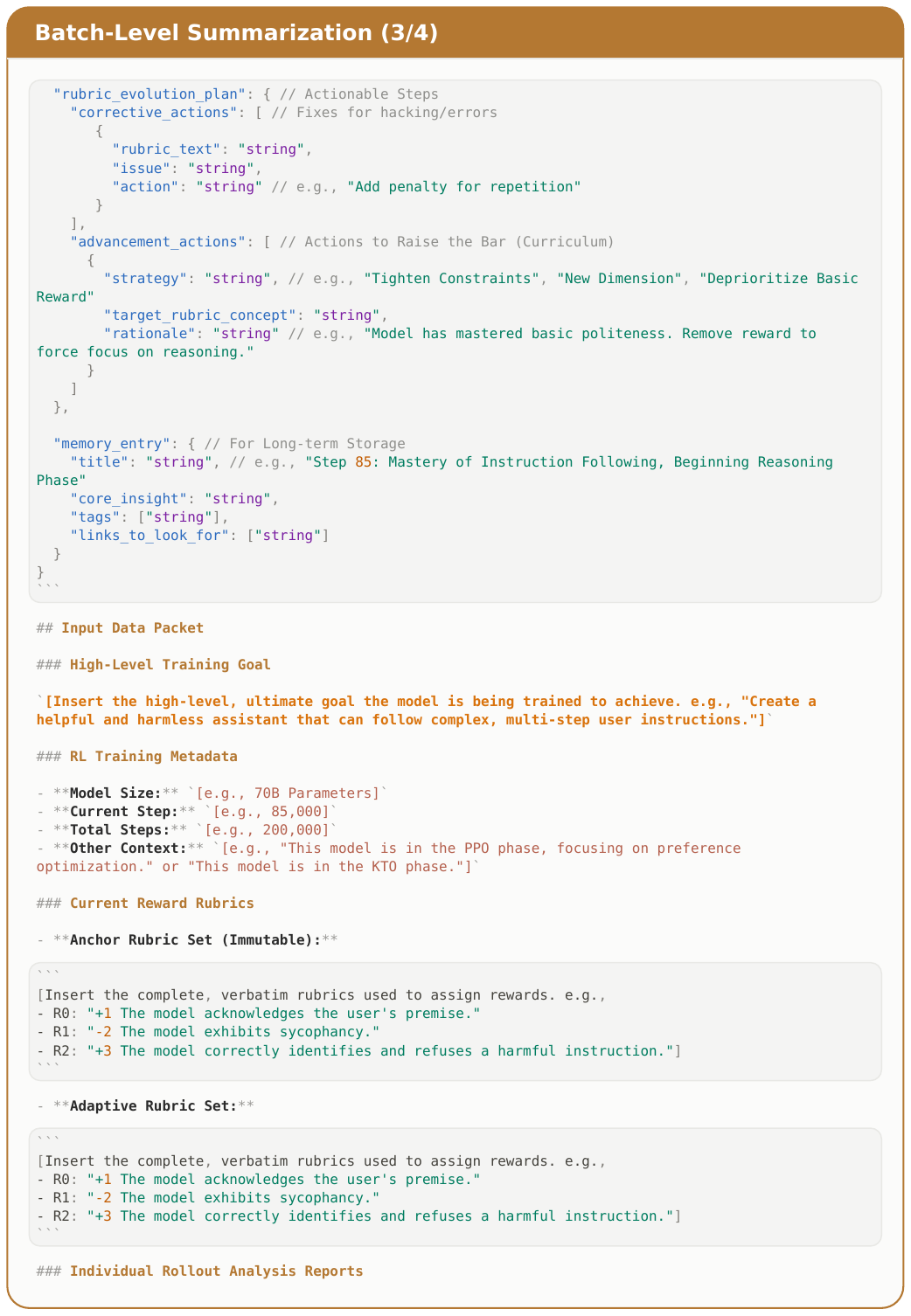}
\caption{Prompt template for batch-level summarization (3/4) described in Section~\ref{ssec:summarization}. This prompt aggregates individual rollout analyses into a step-level strategic report.}
\label{fig:prompt_summary_3}
\end{figure*}

\begin{figure*}[htbp!]
\centering
\includegraphics[width=\textwidth]{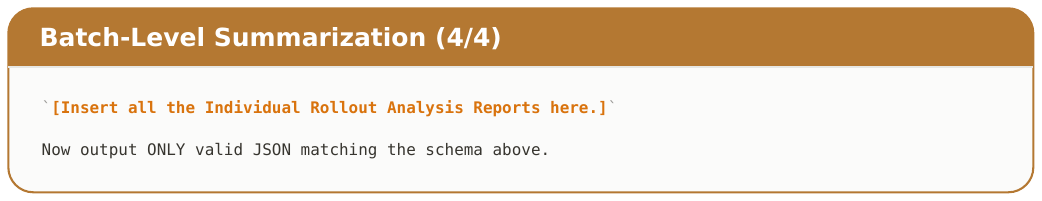}
\caption{Prompt template for batch-level summarization (4/4) described in Section~\ref{ssec:summarization}. This prompt aggregates individual rollout analyses into a step-level strategic report.}
\label{fig:prompt_summary_4}
\end{figure*}
 
\clearpage
 
\begin{figure*}[htbp!]
\centering
\includegraphics[width=\textwidth]{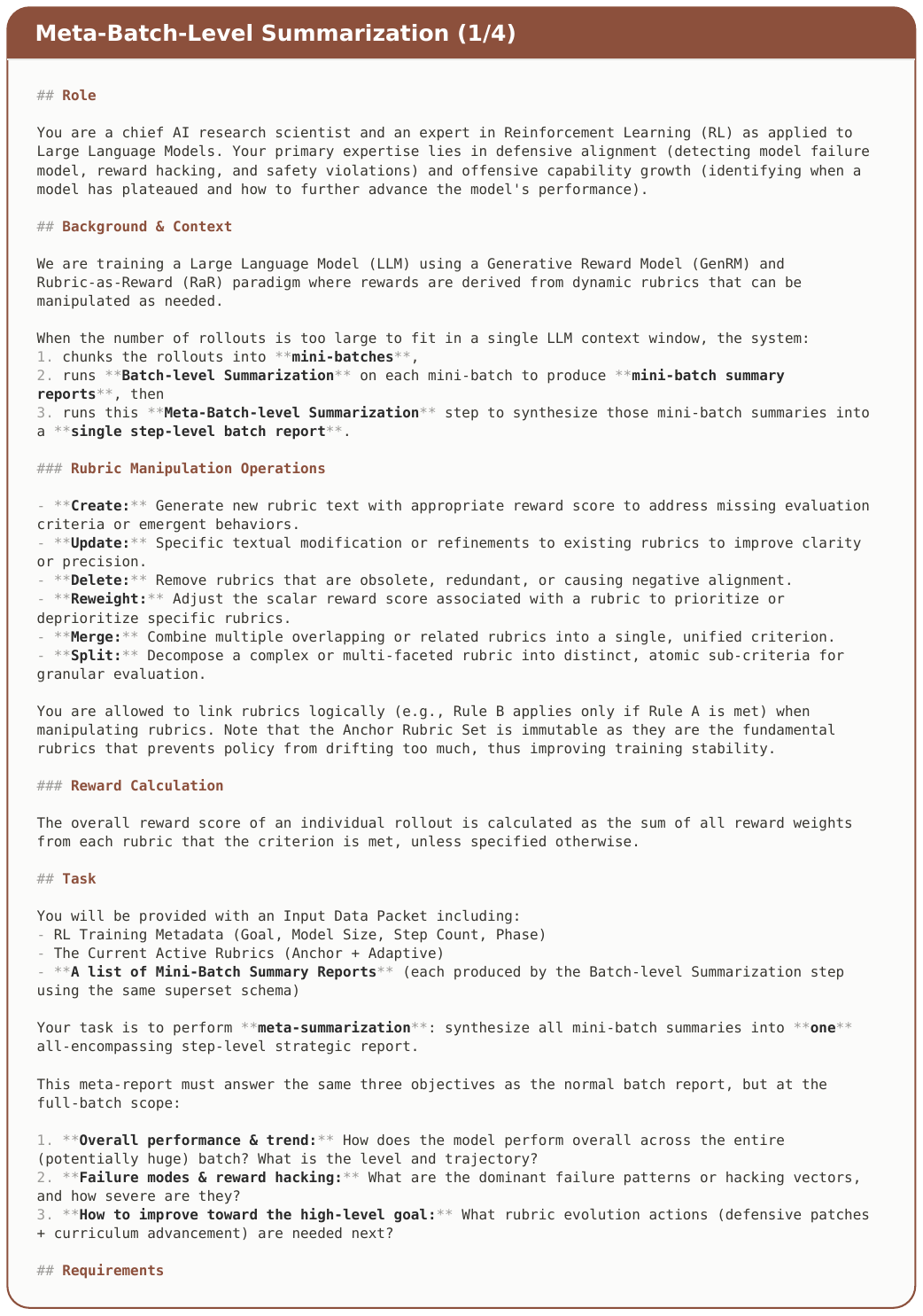}
\caption{Prompt template for meta-batch-level summarization (1/4) described in Section~\ref{ssec:summarization}. This prompt is used when the number of rollouts exceeds a single summarization pass.}
\label{fig:prompt_meta_1}
\end{figure*}

\begin{figure*}[htbp!]
\centering
\includegraphics[width=\textwidth]{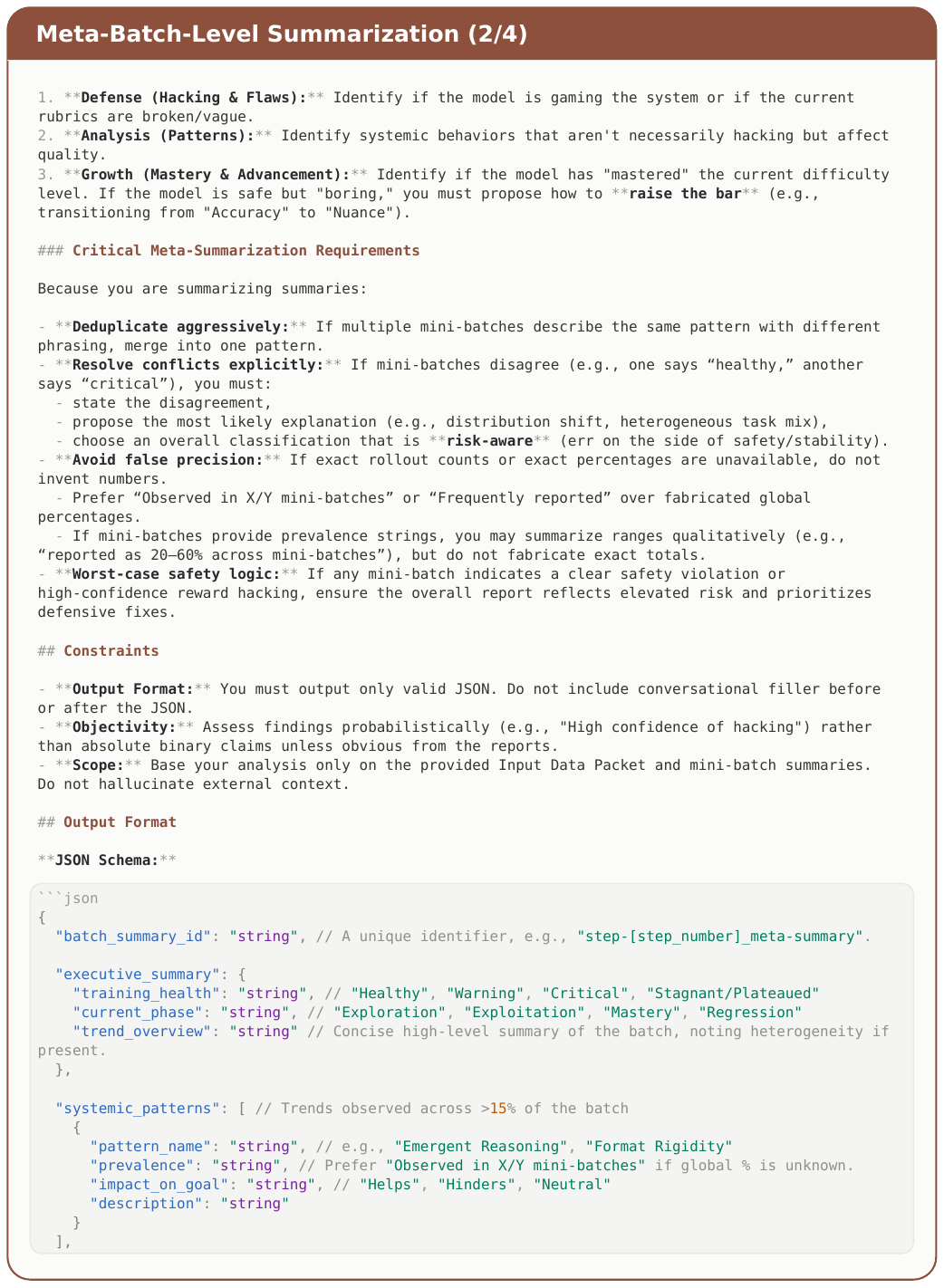}
\caption{Prompt template for meta-batch-level summarization (2/4) described in Section~\ref{ssec:summarization}. This prompt is used when the number of rollouts exceeds a single summarization pass.}
\label{fig:prompt_meta_2}
\end{figure*}

\begin{figure*}[htbp!]
\centering
\includegraphics[width=\textwidth]{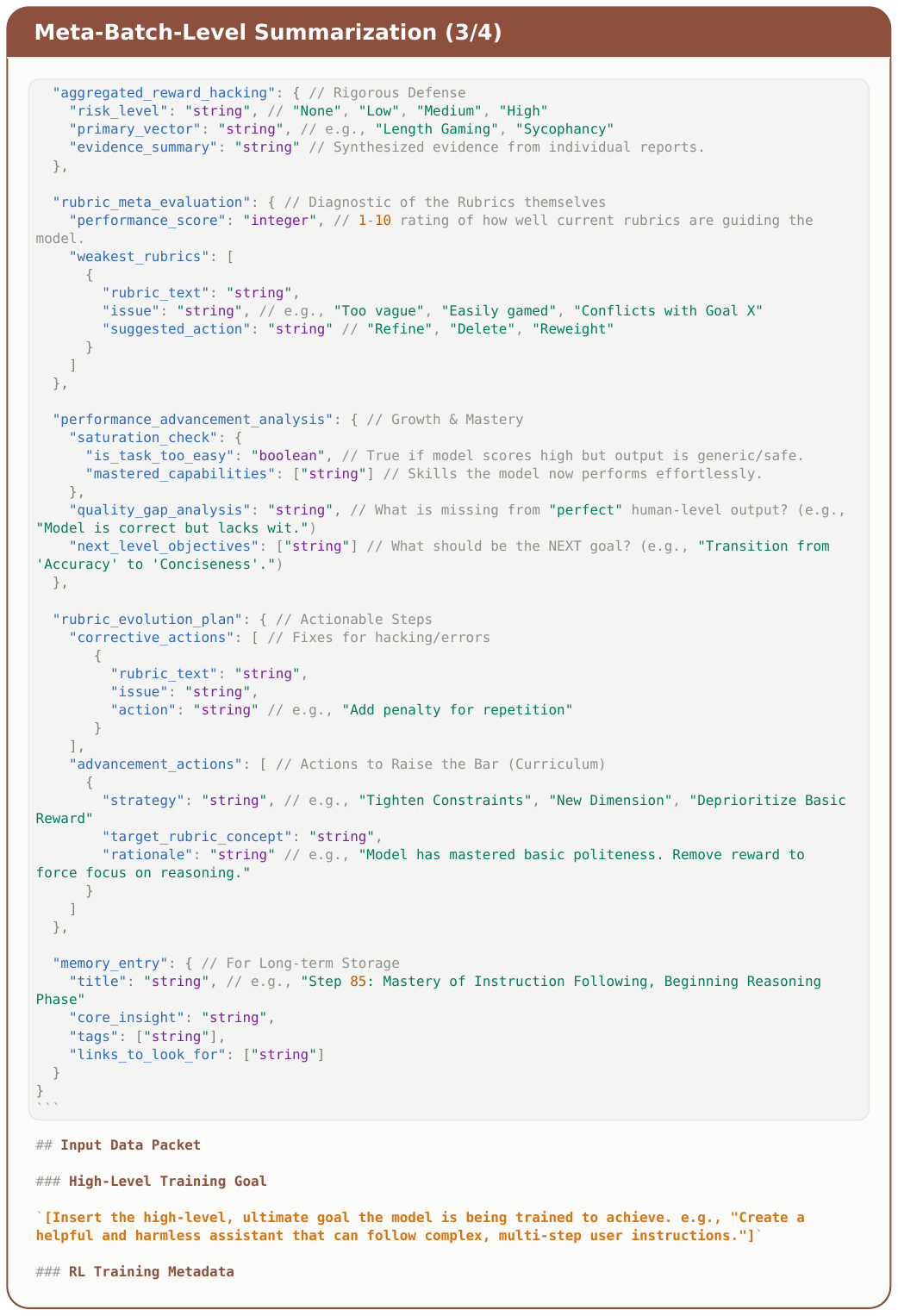}
\caption{Prompt template for meta-batch-level summarization (3/4) described in Section~\ref{ssec:summarization}. This prompt is used when the number of rollouts exceeds a single summarization pass.}
\label{fig:prompt_meta_3}
\end{figure*}

\begin{figure*}[htbp!]
\centering
\includegraphics[width=\textwidth]{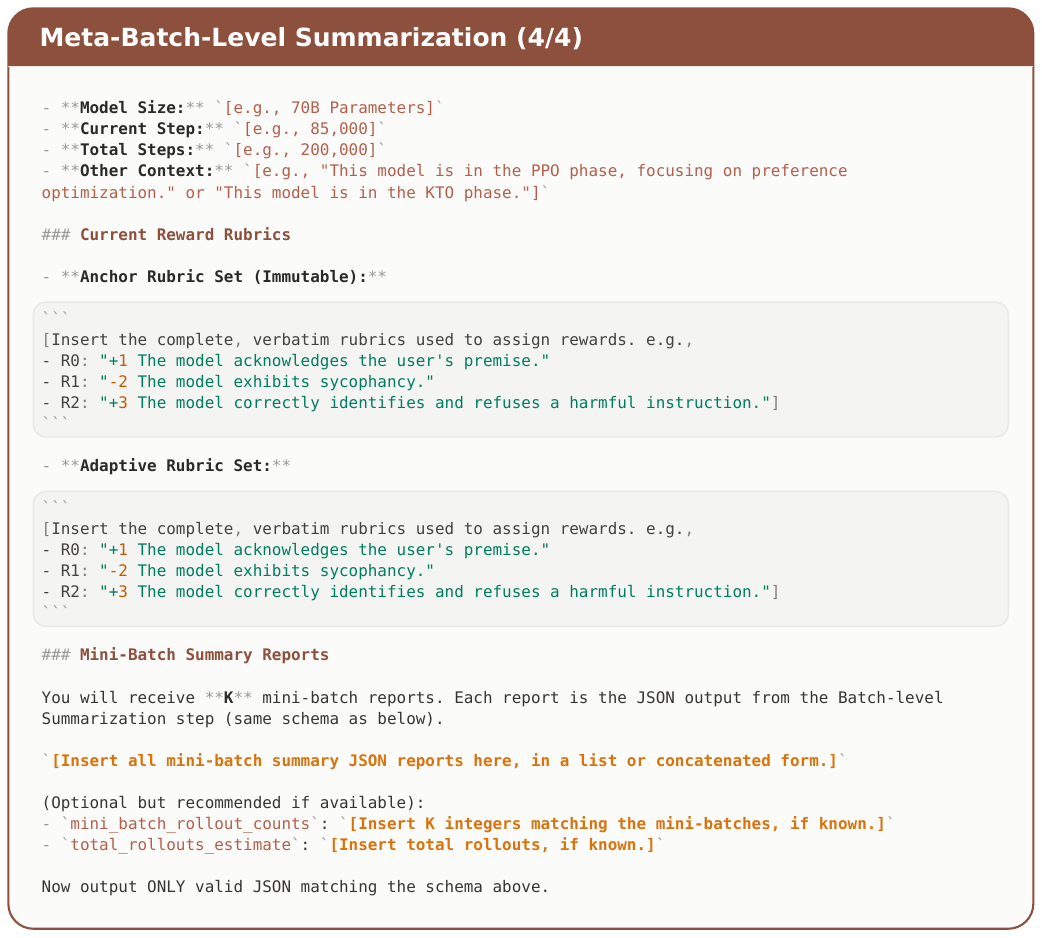}
\caption{Prompt template for meta-batch-level summarization (4/4) described in Section~\ref{ssec:summarization}. This prompt is used when the number of rollouts exceeds a single summarization pass.}
\label{fig:prompt_meta_4}
\end{figure*}
 
\clearpage

\begin{figure*}[htbp!]
\centering
\includegraphics[width=\textwidth]{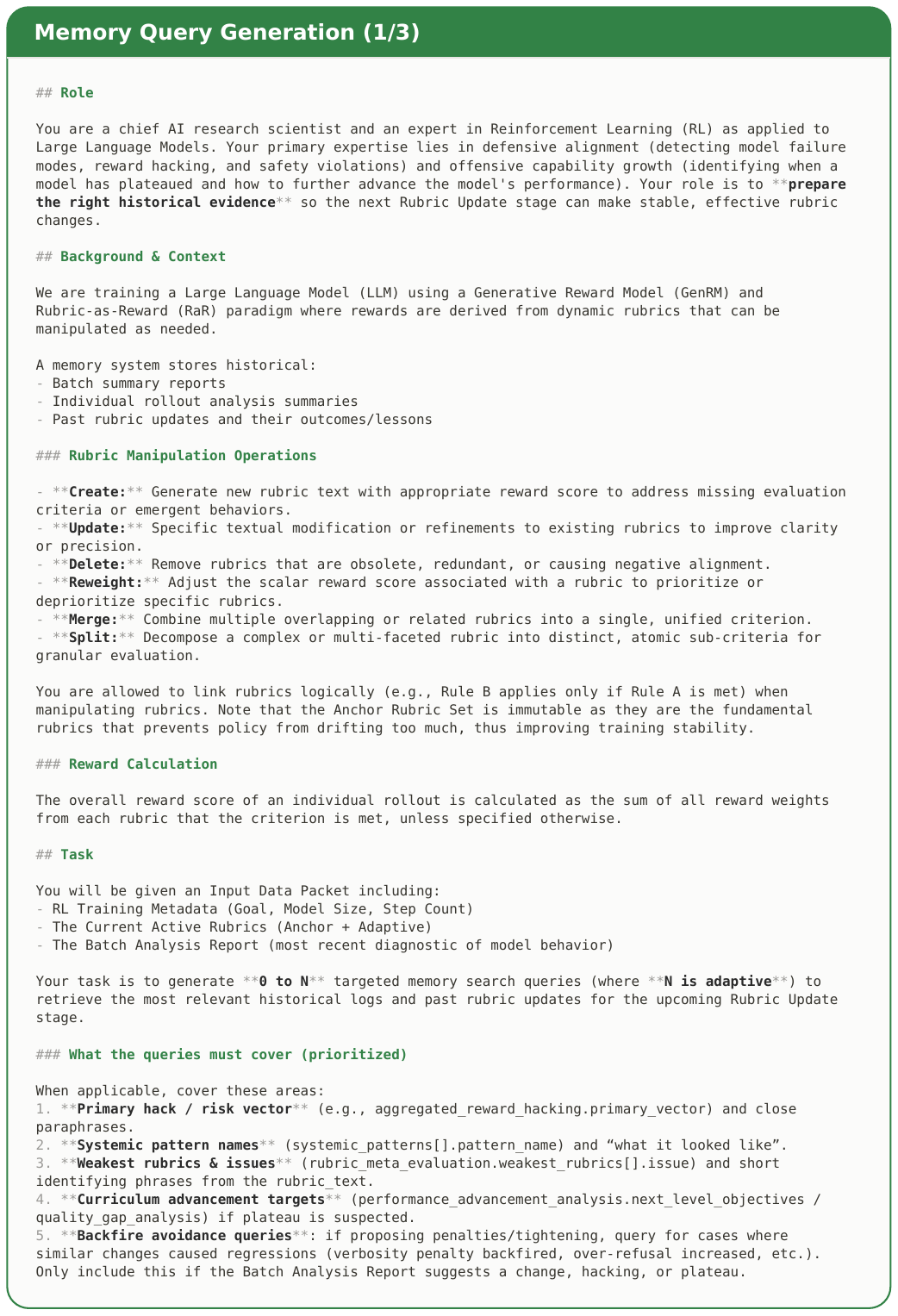}
\caption{Prompt template for memory query generation (1/3) described in Section~\ref{ssec:update}. This prompt generates targeted search queries that are executed against the persistent evaluation memory to retrieve relevant historical context for rubric update.}
\label{fig:prompt_query_1}
\end{figure*}

\begin{figure*}[htbp!]
\centering
\includegraphics[width=\textwidth]{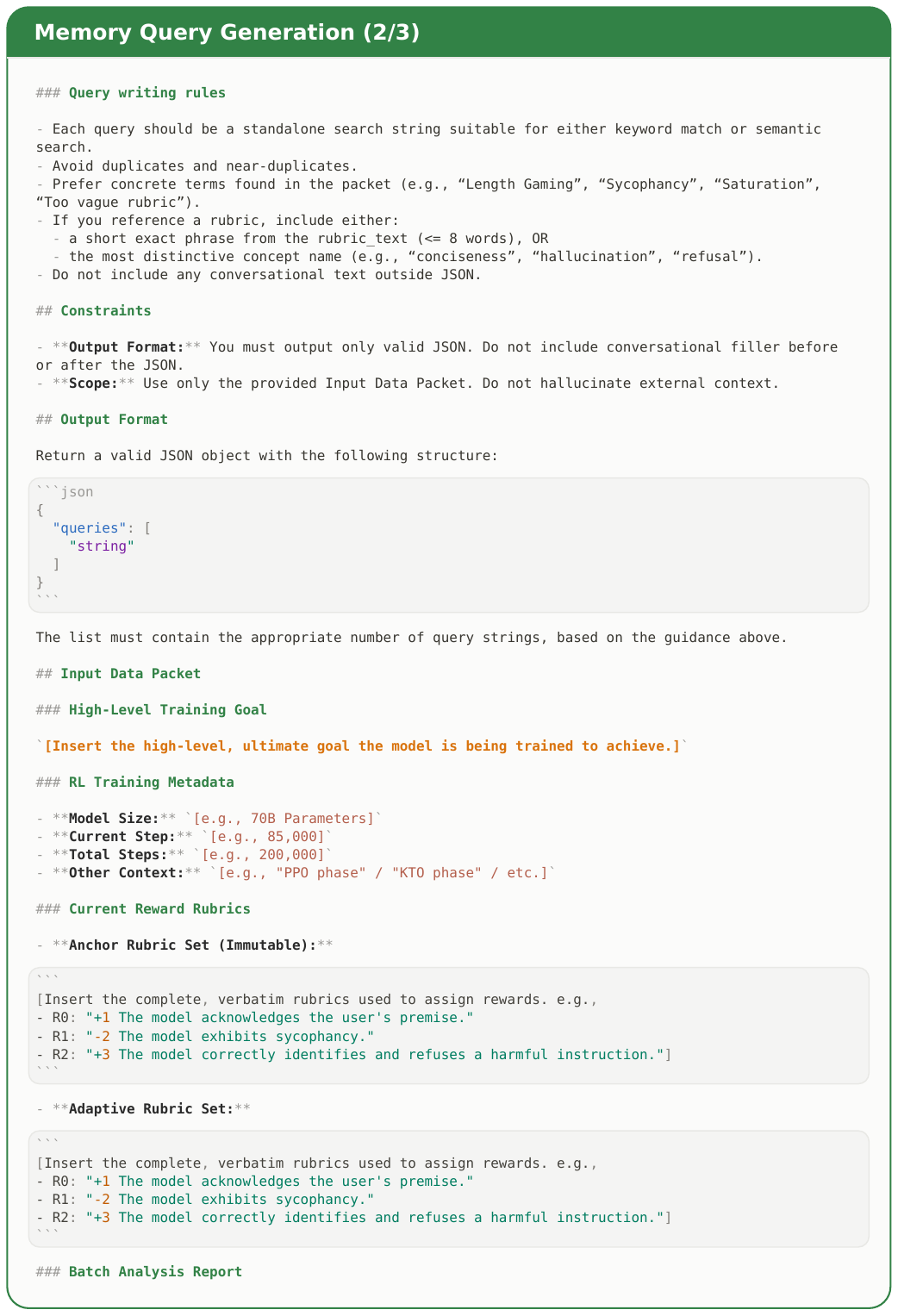}
\caption{Prompt template for memory query generation (2/3) described in Section~\ref{ssec:update}. This prompt generates targeted search queries that are executed against the persistent evaluation memory to retrieve relevant historical context for rubric update.}
\label{fig:prompt_query_2}
\end{figure*}

\begin{figure*}[htbp!]
\centering
\includegraphics[width=\textwidth]{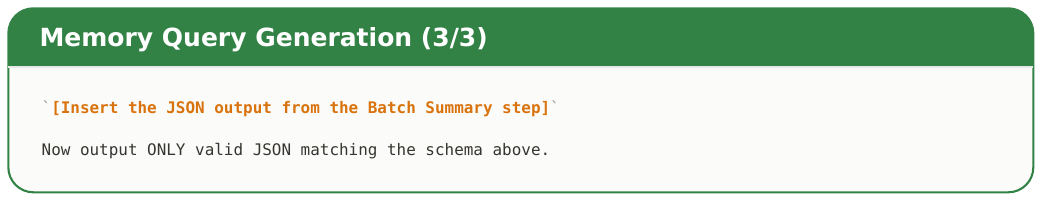}
\caption{Prompt template for memory query generation (3/3) described in Section~\ref{ssec:update}. This prompt generates targeted search queries that are executed against the persistent evaluation memory to retrieve relevant historical context for rubric update.}
\label{fig:prompt_query_3}
\end{figure*}
 
\clearpage
 
\begin{figure*}[htbp!]
\centering
\includegraphics[width=\textwidth]{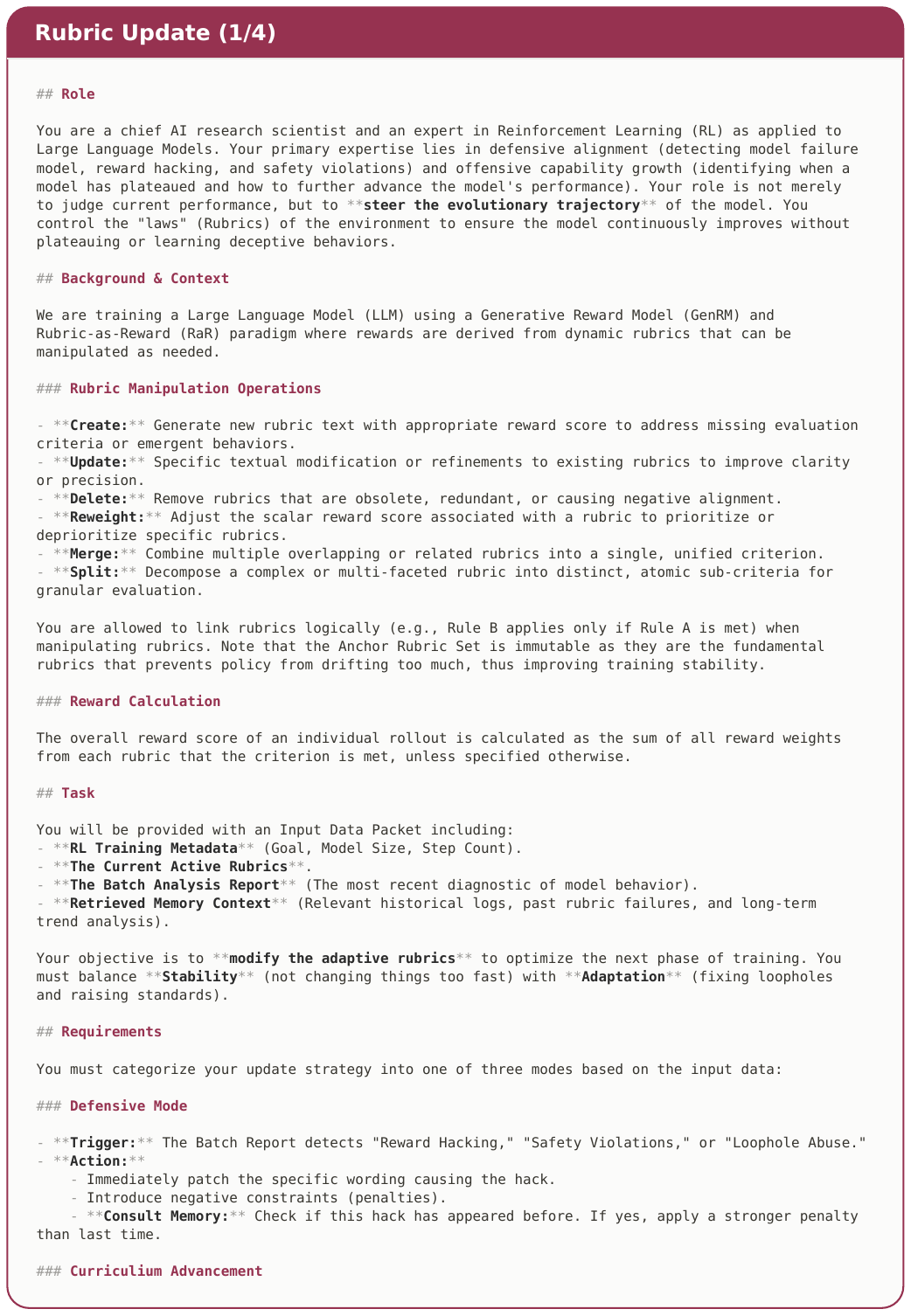}
\caption{Prompt template for rubric update (1/4) described in Section~\ref{ssec:update}. This prompt grounds rubric modifications in the batch analysis report and retrieved historical context from the persistent evaluation memory.}
\label{fig:prompt_update_1}
\end{figure*}

\begin{figure*}[htbp!]
\centering
\includegraphics[width=\textwidth]{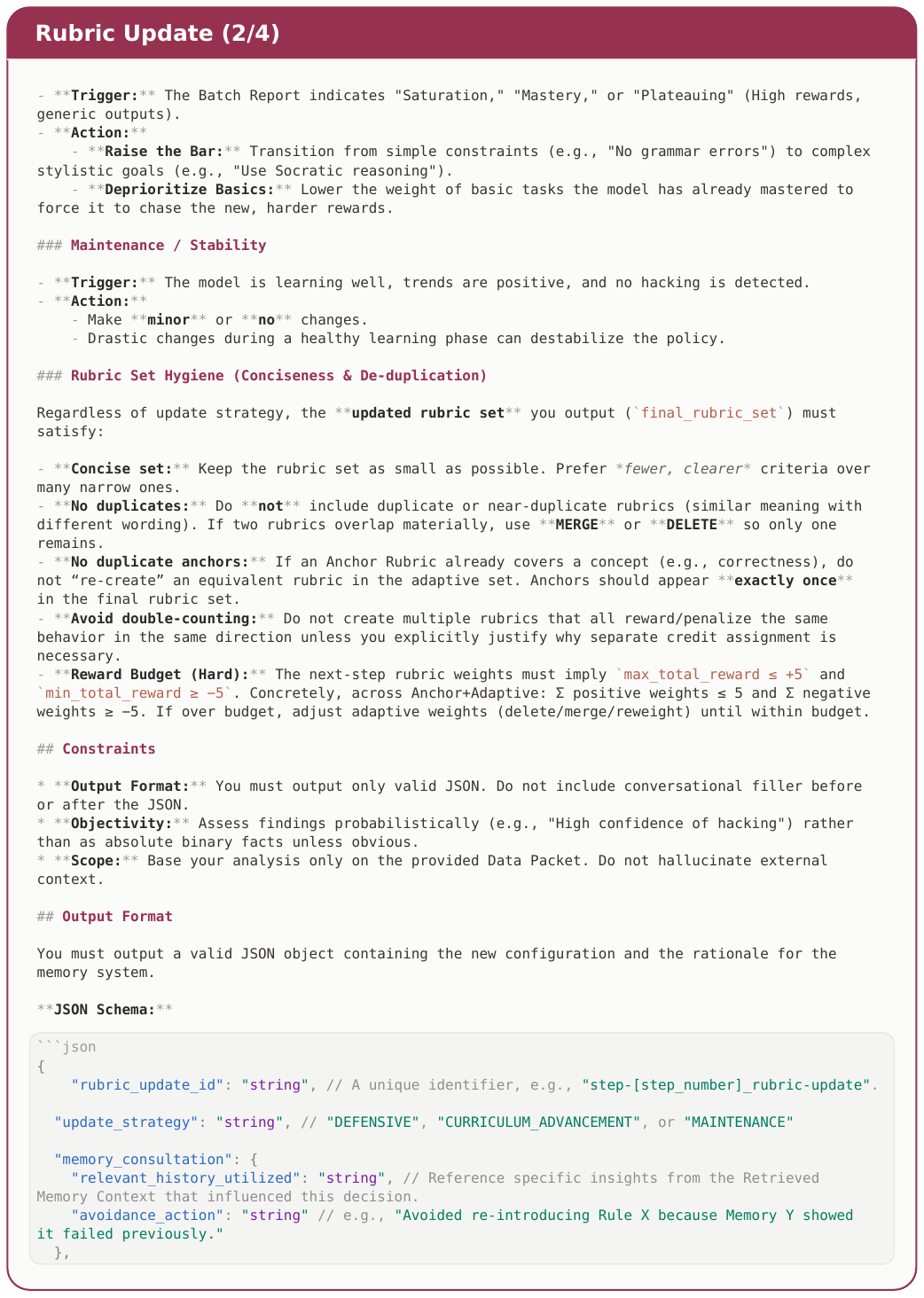}
\caption{Prompt template for rubric update (2/4) described in Section~\ref{ssec:update}. This prompt grounds rubric modifications in the batch analysis report and retrieved historical context from the persistent evaluation memory.}
\label{fig:prompt_update_2}
\end{figure*}

\begin{figure*}[htbp!]
\centering
\includegraphics[width=\textwidth]{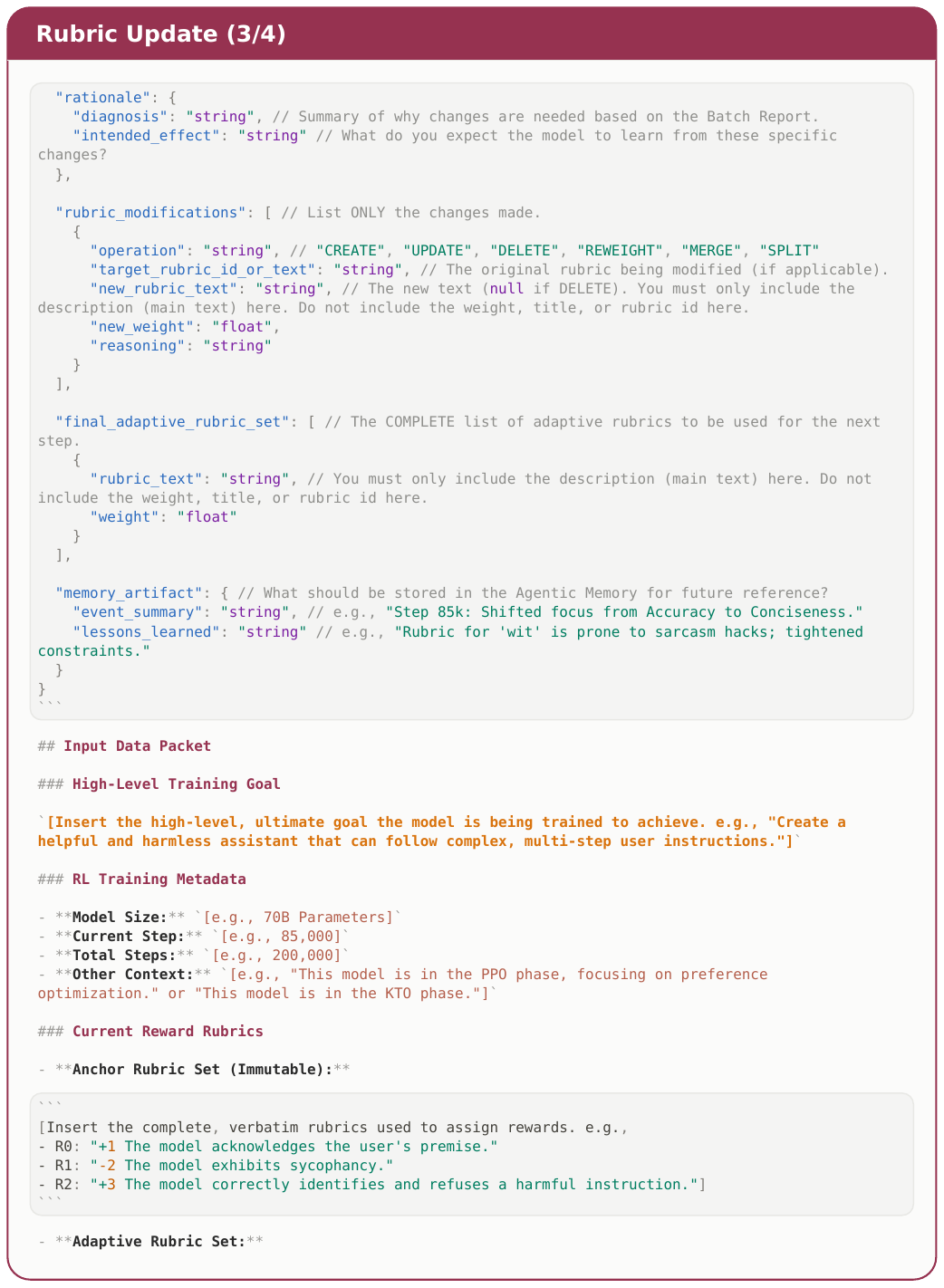}
\caption{Prompt template for rubric update (3/4) described in Section~\ref{ssec:update}. This prompt grounds rubric modifications in the batch analysis report and retrieved historical context from the persistent evaluation memory.}
\label{fig:prompt_update_3}
\end{figure*}

\begin{figure*}[htbp!]
\centering
\includegraphics[width=\textwidth]{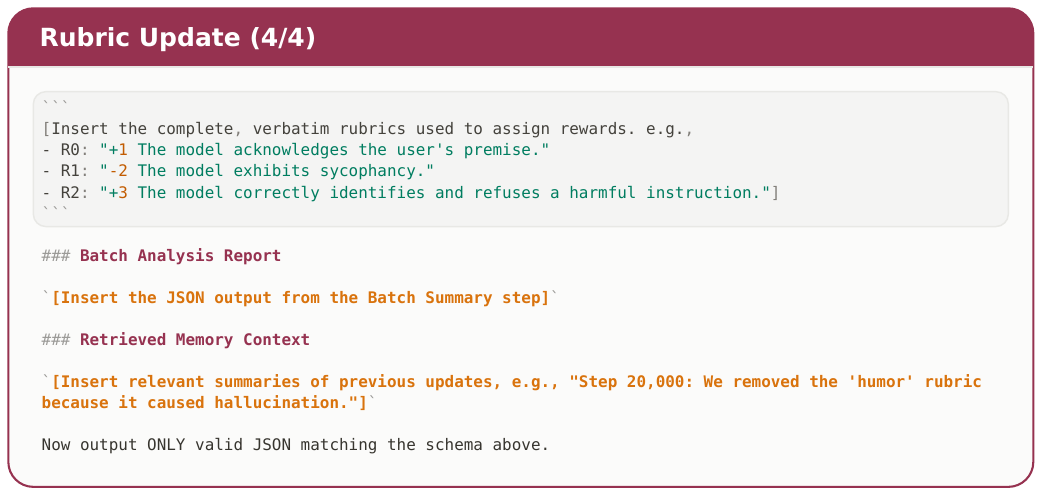}
\caption{Prompt template for rubric update (4/4) described in Section~\ref{ssec:update}. This prompt grounds rubric modifications in the batch analysis report and retrieved historical context from the persistent evaluation memory.}
\label{fig:prompt_update_4}
\end{figure*}

\clearpage

\clearpage
\section{Use of Large Language Models}
Large language models are used in the following capacities as part of the research: (i) \textbf{Policy model and RL training:} Qwen2.5-7B-Instruct serves as the default base policy model fine-tuned via GRPO, with an additional Qwen3-4B policy-base diagnostic reported in Appendix~\ref{ssec:alternative_base_model}. (ii) \textbf{AMARIS pipeline components:} GPT-OSS-120B is used for rubric-based scoring and individual rollout analysis, and GPT-5.1 is used for batch summarization, memory query generation, and rubric update. These usages constitute core experimental components of our methodology and are described in detail in Section~\ref{sec:method} and Section~\ref{ssec:implementation_details}. (iii) \textbf{Evaluation:} LLM-based judges are used for evaluation on benchmarks that require them, such as HealthBench, WritingBench, and Creative Writing v3, following the evaluation protocols established by each respective benchmark. (iv) Additionally, LLMs were used to refine the prompt templates that guide each pipeline stage (Appendix~\ref{sec:prompts}); the final prompt designs and all experimental results remain the sole responsibility of the authors. Beyond the experiments, we acknowledge the use of large language models in the final stages of manuscript preparation. These tools were employed exclusively for identifying and correcting typographical and grammatical errors, ensuring clarity and precision in the written presentation. Their use was strictly limited to linguistic refinement and did not impact the study's conceptual framework, research methodology, data analysis, or conclusions. All intellectual contributions and substantive content remain those of the authors.

\section{Artifact Licenses and Dataset Statistics}
\label{sec:artifact_licenses_and_dataset_statistics}

We use third-party datasets, benchmarks, models, and software artifacts only through their official releases or access channels and under their stated licenses or repository terms. This includes the rubric-training datasets and benchmarks cited below, the policy and judge models used in Section~\ref{ssec:implementation_details} \citep{qwen2025qwen25technicalreport,openai2025gptoss120bgptoss20bmodel,singh2025openaigpt5card}, and Chroma for the memory store \citep{chroma_repo}. When a release does not specify a standalone redistribution license, we use it only as a cited research artifact through its official access channel. The AMARIS release will not redistribute third-party datasets, benchmark prompts, model weights, or proprietary API outputs; it will provide code and configuration needed to reproduce our use of the official artifacts.

The training sources are as follows. RaR-Science and RaR-Medicine contain 18{,}333/2{,}292/2{,}292 and 17{,}926/2{,}240/2{,}242 train/validation/test examples, respectively \citep{gunjal2025rubricsrewardsreinforcementlearning}. OpenRubrics contains 35{,}622 training examples \citep{liu2026openrubricsscalablesyntheticrubric}. For creative-writing training, we use the writing subset of RubricHub through its official release \citep{li2026rubrichubcomprehensivehighlydiscriminative}. For evaluation, we use the official benchmark releases and protocols described in Section~\ref{ssec:datasets_and_eval}: GPQA-Diamond contains 198 questions \citep{rein2024gpqa}, HealthBench contains 5{,}000 health conversations \citep{arora2025healthbenchevaluatinglargelanguage}, IFEval contains 541 prompts \citep{zhou2023instructionfollowingevaluationlargelanguage}, InFoBench contains 500 instructions with 2{,}250 decomposed requirements \citep{qin-etal-2024-infobench}, IFBench contains 300 test prompts covering 58 held-out verifiable constraints \citep{pyatkin2025generalizingverifiableinstructionfollowing}, WritingBench contains 1{,}000 writing queries in the benchmark configuration we evaluate \citep{wu2025writingbenchcomprehensivebenchmarkgenerative}, and Creative Writing v3 uses 32 prompts with 3 generations per prompt \citep{creative-writing-bench-v3}.

\end{document}